\pdfoutput=1

\documentclass[11pt]{article}

\usepackage[final]{coling}

\usepackage{times}
\usepackage{latexsym}

\usepackage[T1]{fontenc}

\usepackage[utf8]{inputenc}

\usepackage{microtype}

\usepackage{inconsolata}

\usepackage{graphicx}

\usepackage{makecell}
\usepackage[normalem]{ulem}
\usepackage{multirow}
\usepackage{url}
\usepackage{tabularray}
\usepackage{booktabs}
\usepackage{subcaption}
\usepackage{comment}
\title{Extending LLMs to New Languages: A Case Study of Llama and Persian Adaptation}

\author{
 \textbf{Samin Mahdizadeh Sani\textsuperscript{1}},
 \textbf{Pouya Sadeghi\textsuperscript{1}},
 \textbf{Thuy-Trang Vu\textsuperscript{3}},
\\
 \textbf{Yadollah Yaghoobzadeh\textsuperscript{1,2}},
 \textbf{Gholamreza Haffari\textsuperscript{3}}
\\
 \textsuperscript{1}Department of Electrical and Computer Engineering, University of Tehran, Iran \\
 \textsuperscript{2}Tehran Institute for Advanced Studies, Khatam University, Iran \\
 \textsuperscript{3}Department of Data Science \& AI, Monash University, Australia \\
 \small{
   \textsuperscript{1}{\{samin.mehdizadeh, pouya.sadeghi, y.yaghoobzadeh\}@ut.ac.ir}
 }
\hspace{1pt}
 \small{
   \textsuperscript{3}{\{trang.vu1, gholamreza.haffari\}@monash.edu}
 }
}

\begin{document}
\maketitle

\begin{abstract}
Large language models (LLMs) have made great progress in classification and text generation tasks. However, they are mainly trained on English data and often struggle with low-resource languages. In this study, we explore adding a new language, i.e., Persian, to Llama (a model with a limited understanding of Persian) using parameter-efficient fine-tuning. We employ a multi-stage approach involving pretraining on monolingual Persian data, aligning representations through bilingual pretraining and instruction datasets, and instruction-tuning with task-specific datasets. We evaluate the model's performance at each stage on generation and classification tasks. Our findings suggest that incorporating the Persian language, through bilingual data alignment, can enhance classification accuracy for Persian tasks, with no adverse impact and sometimes even improvements on English tasks. Additionally, the results highlight the model's initial strength as a critical factor when working with limited training data, with cross-lingual alignment offering minimal benefits for the low-resource language. Knowledge transfer from English to Persian has a marginal effect, primarily benefiting simple classification tasks.\footnote{Code is publicly available at: \href{https://github.com/samin-mehdizadeh/llama-persian-adaptation.git}{https://github.com/samin-mehdizadeh/llama-persian-adaptation.git}}

\end{abstract}

\section{Introduction}
The emergence of large language models (LLMs) has transformed natural language processing (NLP), leading to significant progress in various applications like machine translation, text generation, and sentiment analysis. Models such as GPT-3 \cite{GPT3}, GPT-4 \cite{openai2024gpt4}, and open-source alternatives like Llama-2  \cite{touvron2023Llama2}, Llama-3 \cite{Llama3} and Mistral \cite{jiang2023mistral} have shown impressive abilities in understanding and generating human language. However, these advancements have mostly centered around English and other widely spoken languages, leaving less-resourced languages behind.

In today's connected world, supporting multiple languages in a single model is key to breaking language barriers and making technology accessible to everyone, including speakers of less common languages. Multilingual models help expand the reach of language technologies while having their challenges. Limited data for some languages, keeping performance consistent across languages, and preserving the model's core abilities while adding new languages are the existing challenges \cite{muennighoff-etal-2023-crosslingual, qi-etal-2023-cross, vu-etal-2022-overcoming}. Overcoming these challenges is essential for creating inclusive global language technologies.

Developing multilingual large language models (MLLMs) has become crucial to address these challenges. Models such as XLM-R \cite{XLM_R2019}, Qwen \cite{qwen}, GPT-3 \cite{GPT3}, GPT-4 \cite{openai2024gpt4}, and Llama-3 \cite{Llama3} leverage large multilingual datasets to enhance performance across different languages. However, problems like language imbalance persist, with high-resource languages dominating the training data, resulting in less effective outcomes for low-resource languages.

In this work, we explore cross-lingual adaptation of an LLM, i.e., Llama-2 , focusing on Persian as the target language. Persian, an Indo-European language with a non-Latin script, presents unique challenges due to its linguistic distance from English, the language most LLMs are primarily trained on. Despite its rich literary history and widespread use, Persian has not fully benefited from advancements in LLMs, largely due to limited annotated data and research efforts \cite{benchmarking_farsival}. This makes Persian an ideal case study for testing cross-lingual adaptation. 
We evaluate several models, ranging from a fully English-trained model fine-tuned on Persian data to models with varying degrees of alignment, such as freezing most parameters, and models with additional pre-training combined with Low-Rank Adaptation (LoRA) \cite{hu2021lora}. These configurations create a spectrum of models with different levels of understanding of Persian. Our goal was to examine the differences in performance across these models and assess how well knowledge transfer from English to Persian occurred.

For this study, we use two datasets: one bilingual dataset with parallel data in English and Persian, and another monolingual dataset entirely in Persian. The bilingual dataset allows us to perform cross-lingual training, helping the model learn from aligned English-Persian data. In contrast, the Persian-only dataset is used to fine-tune the model solely on Persian tasks, providing insights into how well the model performs with exclusive exposure to the target language. 

Although there is work adapting LoRA for Persian \cite{abbasi2023persianLlama,rostami2024persianmind}, they typically focus on a single model for this language. Our work, however, explores different settings for model pre-training and instruction-tuning across both classification and generation tasks, highlighting the effort required to adapt models for specific tasks. 

Our findings indicate that for classification tasks, aligning models with bilingual data is sufficient, and further pre-training on monolingual data is not crucial. However, when dealing with limited data or generation tasks, further pre-training proves beneficial and improves performance, as generation requires a deeper understanding of linguistic structures. Moreover, evaluations of several models that initially support the Persian language reveal the weaknesses of Llama models for this language, leaving it behind Gemma~\cite{team2024gemma} and Qwen models. However, the comparison of our instruction-tuned model with state-of-the-art systems underscores the potential of targeted fine-tuning. Our model achieves performance comparable to these advanced models across most tasks. Notably, for simple yet unseen tasks like sentiment analysis, the model demonstrates an ability to generalize patterns from other instructions. Nevertheless, for more complex tasks that challenge the model’s knowledge, it struggles to perform effectively without explicit instructions.

\section{Related Work}

\paragraph{Cross-lingual Transfer.}
The rapid advancement of large-scale language models, such as GPT-3 \cite{GPT3} and GPT-4 \cite{openai2024gpt4}, has significantly enhanced the capabilities of natural language processing (NLP). Moreover, Recent contributions from the open-source community, including Llama \cite{touvron2023Llama2} and Mistral \cite{jiang2023mistral}, show that open-source LLMs can now compete effectively with their closed-source counterparts. However, the heavy focus on English limits the flexibility of these models when incorporating new languages, particularly low-resource ones, which may not be initially supported in the training phase. Recent studies have leveraged the proficiency of language models in English to enhance performance on low-resource languages. Some of these studies focus on translating model input into English, demonstrating notable improvements for low-resource languages \cite{upadhayay2023taco,benchmarking_farsival}. Another approach explores transliteration, either incorporating it during the pre-training phase \cite{moosa-etal-2023-transliteration,purkayastha-etal-2023-romanization} or adapting it in the fine-tuning stage \cite{dabre-etal-2022-indicbart,muller-etal-2021-unseen}. Additionally, several works have examined the impact of adding translation instructions alongside target language instructions, showing consistent gains in performance \cite{ranaldi-pucci-2023-english,ranaldi2023empowering,zhu2023extrapolating}. In our study, in addition to fine-tuning models using bilingual English-Persian instructions, we conduct additional experiments to assess cross-lingual transfer between English and Persian. While most previous studies focus on combining translation instructions with target language instructions, we exclude any instructions in the target language (Persian). Instead, we fine-tune the models using only English and translation instructions to evaluate whether the model's proficiency in English and its ability to translate into Persian can eliminate the need for explicit target language instructions, particularly for a linguistically distant language like Persian.

\paragraph{Language Adaptation.} 

Multilingual language models, such as mBERT \cite{pires-etal-2019-multilingual}, XLM-R \cite{XLM_R2019}, GPT-4 \cite{openai2024gpt4}, and Llama-3 \cite{Llama3}, rely on large multilingual datasets to learn linguistic structures. However, they still face challenges like language imbalance, where high-resource languages dominate. Techniques like parameter-tuning and parameter-freezing have been proposed to enhance performance \cite{qin2024multilingual}, though these methods may not always be suitable for extending monolingual models. Research has focused on efficiently adding new languages to LLMs, with some studies investigating ways to mitigate catastrophic forgetting when integrating new languages \cite{csaki2023efficientlyadaptingpretrainedlanguage,alexandrov2024mitigatingcatastrophicforgettinglanguage}. Others aim to improve models for specific languages, such as Arabic \cite{gosal2024bilingualadaptationmonolingualfoundation}, Chinese \cite{cui2023efficient,ji2023towards}, and Persian \cite{abbasi2023persianLlama,rostami2024persianmind}, through vocabulary extension and additional pre-training using parameter-efficient fine-tuning (PEFT). These studies typically focus on a single language model and do not systematically evaluate models across each training phase. The work most similar to ours, \cite{tejaswi2024exploring}, explores design choices like base model selection and vocabulary size for low/mid-resource languages. Nonetheless, it does not emphasize bilingual training during pre-training and instruction tuning and instead performs full fine-tuning, which limits its focus on the constraints of PEFT techniques.

\section{Data}

\begin{figure*}[htbp]
\centering
    \begin{subfigure}{0.47\textwidth}
      \includegraphics[width=\linewidth]{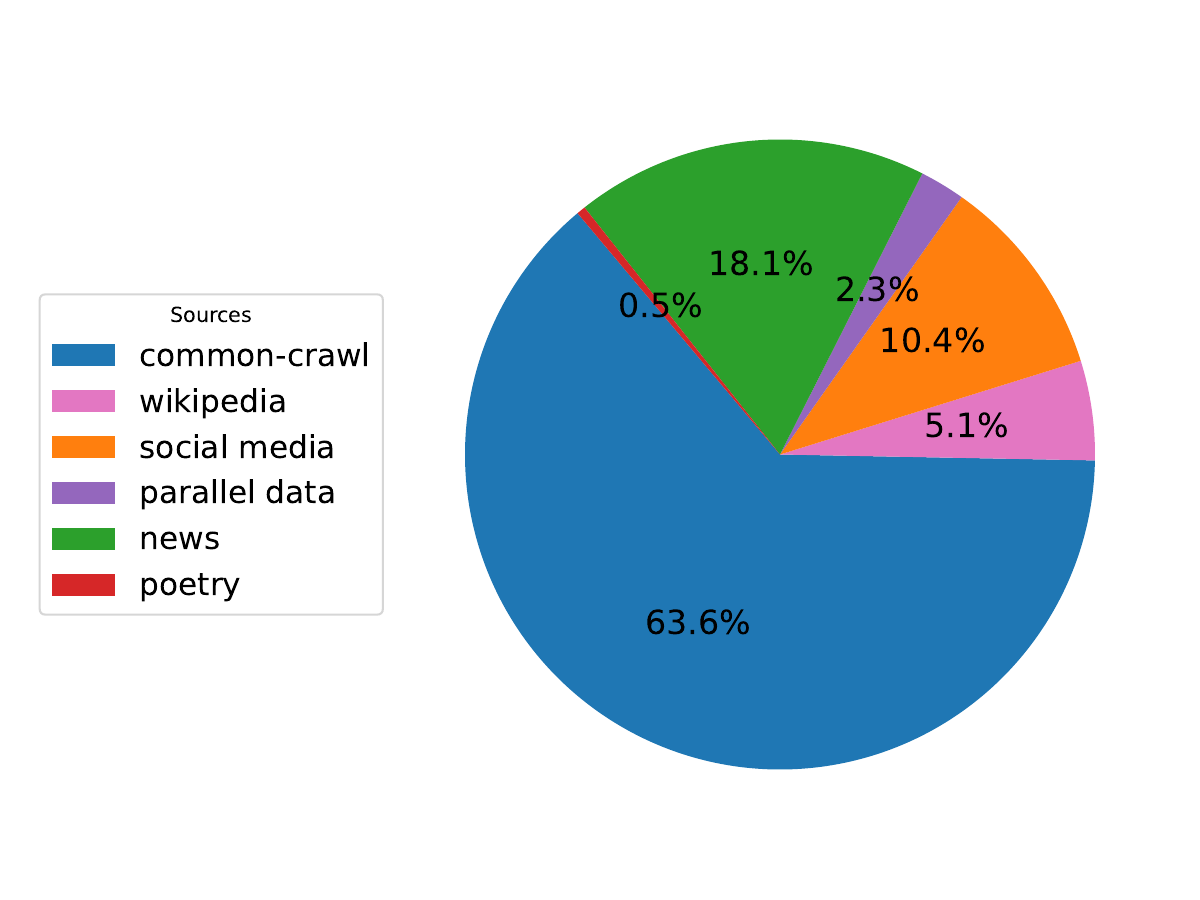}
      \caption{Persian pre-training data sizes: \textbf{\textit{12.96GB}} monolingual and \textbf{\textit{279MB}} parallel.}
      \label{fig:dataset:pre-train}
    \end{subfigure} 
    \hfill
    \begin{subfigure}{0.47\textwidth}
      \includegraphics[width=\linewidth]{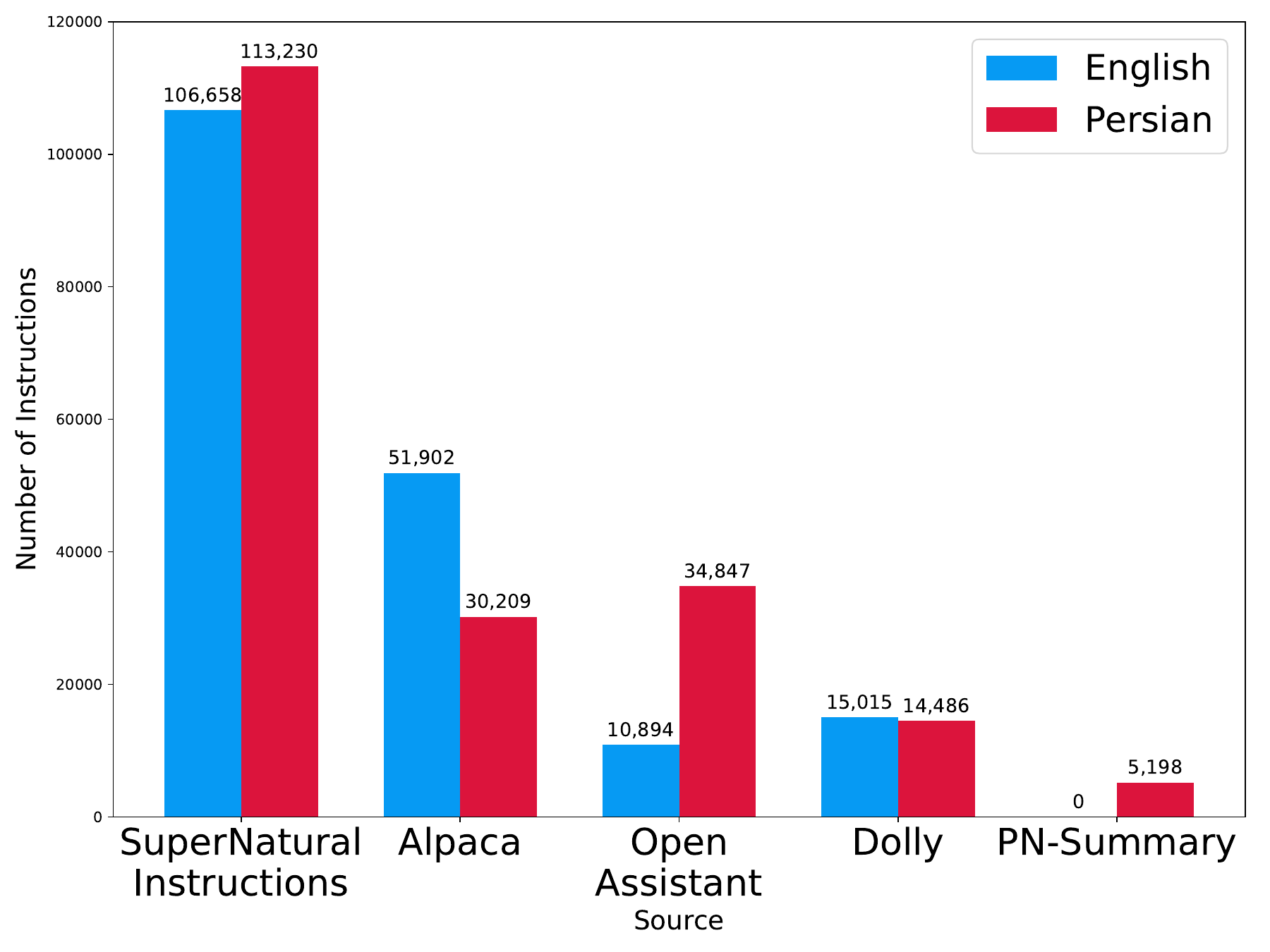}
      \caption{Number of instructions used: \textbf{\textit{184,496}} in English and \textbf{\textit{197,970}} in Persian.}
      \label{fig:dataset:inst}
    \end{subfigure} 
    \caption{Statistics of datasets used for pre-training and instruction-tuning.}
    \label{fig:datasets}
\end{figure*}

Building on the advancements outlined in related work, we collected data for both pre-training and instruction tuning to extend the LLM to Persian. This section introduces the datasets used for pre-training (Figure \ref{fig:dataset:pre-train}) and instruction tuning (Figure \ref{fig:dataset:inst}), utilizing both English and Persian texts.

\subsection{pre-training}\label{corpora}
pre-training data can significantly influence the performance of models. To ensure a wide range of data, we collect the pre-training data from five different categories: news, poems, Wikipedia, Twitter, and data collected by crawling web pages. The data were collected from seven different sources including Leipzig\footnote{\url{https://corpora.uni-leipzig.de/}}, LSCP \cite{abdi-khojasteh-etal-2020-lscp}, Miras \cite{sabeti-etal-2018-mirastext}, Farsi-Poems\footnote{\url{https://github.com/amnghd/Persian_poems_corpus}}, VOA (Voice of America) \footnote{\url{https://jon.dehdari.org/corpora/}}, Wikipedia\footnote{\url{https://github.com/Text-Mining/Persian-Wikipedia-Corpus}} and YJC-news\footnote{\url{https://github.com/mohammadiahmad/persian-dataset}}. 

\begin{table}[!h]
\centering
\resizebox{\linewidth}{!}{%
\begingroup
\setlength{\tabcolsep}{5pt} 
\renewcommand{\arraystretch}{1.5} 
\begin{tabular}{l|l|r|r} 
\toprule
\textit{\textbf{Perian source}}                                & \textit{\textbf{Type}} & \textit{\textbf{Original}} & \textit{\textbf{pre-processed}}  \\ 
\toprule
\textit{Leipzig}                                    & news/commoncrawl                   & 424.36 MB                  & 414.93 MB                       \\ 
\textit{LSCP}                                       & twitter                & 2.73 GB                    & 1.27 GB                         \\ 
\textit{MirasText}                                            & commoncrawl            & 14.62 GB                   & 7.34 GB                         \\ 
\textit{FarsiPoems}                                                & poem                   & 60.72 MB                   & 55.59 MB                        \\ 
\textit{VoaPersian}                                           & news                   & 66.48 MB                   & 59.13 MB                        \\ 
\textit{Wikipedia}                                            & wikipedia              & 845.10 MB                  & 622.59 MB                       \\ 
\textit{YJCNews}                                             & news                   & 2.85 GB                    & 2.15 GB                         \\ 
\midrule
\multicolumn{2}{c|}{\textit{\textbf{Total Data size}}} & {\textbf{21.6GB}} & {\textbf{12.96GB}}                                                 \\
\bottomrule
\end{tabular}
\endgroup
}
\caption{
pre-training data before and after pre-processing.
}
\label{tab:pre-trained-data}
\end{table}

All of the documents have been pre-processed. Our cleaning pipeline follows the procedure introduced in \cite{raffel2020exploring}. We first extract sentences from each document and remove those with fewer than five words. Additionally, sentences containing special keywords from Persian web pages or characters indicating a piece of code were removed. Finally, sentences with a probability of being Persian of less than 70\% were removed. This threshold was chosen to include some English texts alongside Persian texts, ensuring the data was not entirely in Persian. It is worth noting that only unique sentences were kept, and duplicates were removed. Table~\ref{tab:pre-trained-data} indicates our collected data for the pre-training.

\vspace{4mm}
In addition to Persian datasets, we also leverage parallel English-Persian corpora to align newly added Persian embeddings with English ones. The parallel corpora were collected from MIZAN \cite{Kashefi2018MIZNA}, TEP \cite{Pilehvar2011TEPTE}, and PEPC \cite{karimi-etal-2018-extracting}.  Table ~\ref{tab:parallel-corp} includes the details of our parallel datasets.

\begin{table}[!h]
\centering
\resizebox{0.7 \linewidth}{!}{%
\begingroup
\begin{tabular}{lrr} 
\toprule
\textbf{\textit{Source}}             & \textbf{\textit{Size (En)}}             & \textbf{\textit{Size (Fa)}}       \\ 
\toprule
\multirow{1}{*}{\textit{Mizan}}               & 62.55 MB     & 106.97 MB                \\ 

\multirow{1}{*}{\textit{TEP}}                 & 20.11 MB             & 32.40 MB                 \\ 
\multirow{1}{*}{\textit{PEPC}} & 24.55 MB   & 36.42 MB                 \\
\midrule
\textbf{\textit{Total}}                    & \textbf{\textit{105.38 MB}}                    & \textbf{\textit{173.55 MB}}  \\
\bottomrule
\end{tabular}
\endgroup
}
\caption{
The parallel English-Persian datasets after cleaning and deduplication.
}
\label{tab:parallel-corp}
\end{table}

\subsection{Instruction Tuning}\label{sec:inst-data}
In order to enable the model to follow instructions, we also perform instruction tuning. According to \citealt{wang2023far}, the more diverse the training data in the instruction tuning phase, the stronger and more generalizable the resulting model will be. Unfortunately, the variety of instructions for the Persian language is very limited. Consequently, in addition to Persian instructions, we also utilized the English ones. For English instructions, the data were compiled from Alpaca \cite{alpaca}, Dolly \cite{DatabricksBlog2023DollyV1}, OpenAssistant \cite{kopf2024openassistant}, and Super Natural Instructions \cite{supernaturalinstructions}. For Persian instructions, translated and cleaned versions of these corpora, as well as PN-Summary \cite{farahani2021leveraging} were used. Table~\ref{tab:all-inst} shows the number of instructions used in both languages.

\begin{table}[!ht]
\centering
\resizebox{\linewidth}{!}{%
\begingroup
\setlength{\tabcolsep}{10pt} 
\renewcommand{\arraystretch}{1.5} 
\begin{tabular}{l|r|r} 
\toprule
\textbf{Source}      & \textbf{Count (En)} & \textbf{Count (Fa)}    \\ 
\toprule
Alpaca               & 51,902            & 30,209            \\
Dolly                & 15,015            & 14,486            \\
Super-Natural-Instructions & 106,658           & 113,230           \\
OpenAssistant        & 10,894            & 34,847            \\ 
PN-Summary           & -                 & 5,198             \\ 
\midrule
\textbf{All}         & \textbf{184,496}  & \textbf{197,970}  \\ 
\bottomrule
\end{tabular}
\endgroup
}
\caption{
Number of instructions in both English and Persian languages, used for instruction tuning.
}
\label{tab:all-inst}
\end{table}

\section{Training Details}\label{sec:train}
Our goal is to train a model that can comprehend the Persian language with limited training data and transfer the knowledge it has in English to answer Persian tasks. To achieve this, we leverage the Llama-2-7B model, which performs well in English but does not understand Persian properly. We then extend the model's vocabulary with Persian words, pre-train it with both monolingual and bilingual datasets, and finally, instruction-tune it with instructions in both languages.

\subsection{Vocabulary Expansion} 
To enhance the model's ability in the Persian language and also reduce training and inference time, we first train a SentencePiece \cite{kudo2018sentencepiece} model to extract Persian tokens from 4GB of Persian data,  randomly selected from our Persian corpora introduced in Section \ref{corpora}. We then merge the 20,000 new Persian tokens with the 32,000 original model vocabulary tokens. About 10\% of the new tokens overlapped with the original model vocabulary, resulting in a total of 49,816 tokens in the model. Finally, the model embeddings are expanded to include the additional Persian vocabulary.

\subsection{Token Prediction}
We perform two phases to train our model with the objective of the next token prediction. In the first phase, the focus is on learning new Persian representations and aligning them with English embeddings, while the second phase emphasizes Persian text generation.

\subsubsection{Embedding Alignment}
To obtain representations for Persian tokens and align them with the English ones, all transformer layers are frozen, and only the heads and embeddings are trained. The training is done using monolingual and bilingual next-token prediction. In the first step, the Persian corpora are used to improve the embeddings for Persian tokens. In this stage, the model only needs to predict Persian words. In the second step, the model is trained for token prediction in a bilingual manner. First, the model predicts Persian text. Then, it generates English sentences which are translations of the Persian sentences. Throughout this process, the model is penalized based on our parallel data during training.  This approach not only maintains the model's proficiency in English but also aligns the learned representations for Persian with English ones. For this step, we leverage the parallel dataset introduced in section ~\ref{corpora}.

\subsubsection{Text Generation}
\begin{figure}[ht]
	\centerline{\includegraphics[width=0.5\textwidth]{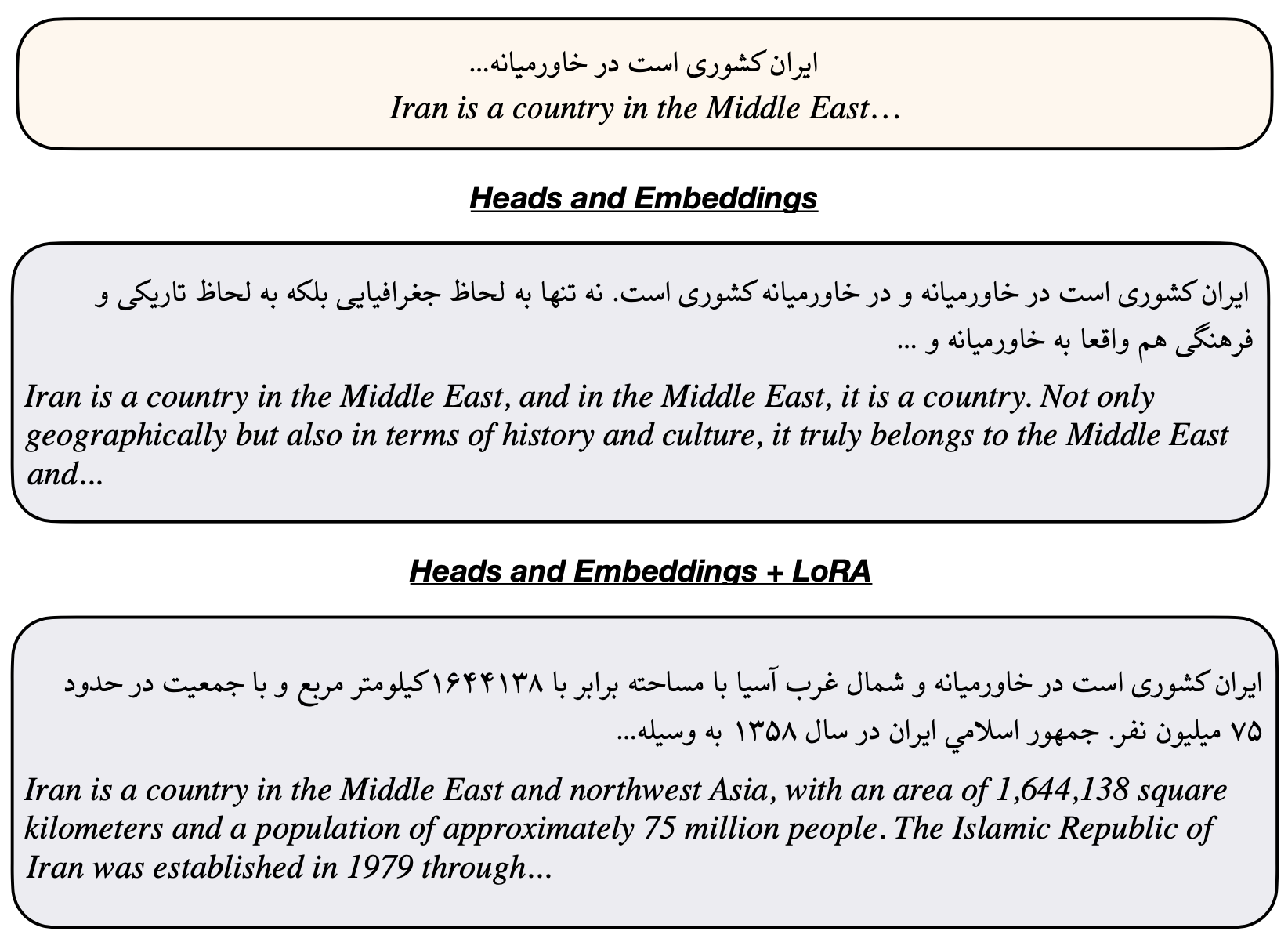}}
	\caption{
Generated text before and after LoRA fine-tuning. Models are tasked with completing the sentence in the orange box. The text generated without LoRA does not follow the expected syntactic and semantic structure.
	}
 \label{fig:generation}
\end{figure}

Training the heads and embeddings alone is not sufficient for generating Persian text. The model obtained from the previous section has limited knowledge about the semantic and syntactic structure of Persian sentences (see Figure \ref{fig:generation}). Therefore, we continue training the model by adding weights from LoRA using a monolingual Persian dataset. At this stage, in addition to fully updating the heads and embedding layers, the LoRA weights are also updated across all layers. We use LoRA to first accelerate the training time and reduce the required resources, and second, to maintain the model's capabilities for the English language by only changing limited weights. Finally, the model is instruction-tuned using both Persian and English instructions to align the two languages. Notably, around 24\% of the instructions are related to $English \leftrightarrow Persian$ translations.

\subsection{Configuration}

Model training and fine-tuning in all stages are performed using eight V100 GPU units. To optimize memory usage, the Zero Redundancy Optimizer \cite{rajbhandari2020zero} was applied, and the weights' gradients, along with the optimizer's variables, were distributed among the processing devices. Moreover, all variables were stored in half-precision floating-point format (fp16). In all stages where the LoRA model was used, the rank was 8, and the value of $\alpha$ was 32 (the weight update coefficient was $\alpha / r = 4$). Table \ref{tab:train-percent} indicates the percentage of trainable parameters in each step.

\begin{table}[!ht]
\begingroup
\setlength{\tabcolsep}{6pt} 
\renewcommand{\arraystretch}{1.7}
\centering
\resizebox{\linewidth}{!}{%
\begin{tabular}{l|r|r|r}
\toprule
\textbf{Parameter}                  & \textbf{Alignment} & \textbf{pre-training}  & \textbf{Instruction-tuning}  \\ 
\toprule
\textit{\textbf{All}}      & 6,884,372,480         & 6,904,360,960 & 6,904,360,960       \\ 
\textit{\textbf{Trainable}} & 408,100,864           & 428,089,344   & 428,089,344         \\ 
\midrule
\textbf{\textbf{Percentage}}         & \textbf{5.93\%}                & \textbf{6.20\%}          & \textbf{6.20\%}                \\
\bottomrule
\end{tabular}
}
\endgroup
\caption{
Percentage of trainable parameters in each stage. pre-training and instruction-tuning are done with LoRA.
}
\label{tab:train-percent}
\end{table}

\section{Evaluations}
Our evaluations include two types of tasks: classification and text generation. The assessments are performed on various models obtained from each phase (see section ~\ref{sec:train}). This section presents the experiments as well as our analysis of different training strategies.

\begin{table*}[ht]
\resizebox{\textwidth}{!}{%
\begin{tabular}{@{}lcccccccc@{}}
\cmidrule(l){2-9}
\multicolumn{1}{c}{}                    & \multicolumn{8}{c}{\textbf{Training Instructions: English + Persian}}                                                                \\ \cmidrule(l){2-9} 
\multicolumn{1}{c}{}                    & \multicolumn{4}{c}{English Tasks}                                           & \multicolumn{4}{c}{Persian Tasks}                      \\ \cmidrule(l){2-9} 
\multicolumn{1}{c}{}                    & Random & Llama-2        & Em-aligned    & \multicolumn{1}{c||}{Fa-pretrained} & Random & Llama-2        & Em-aligned    & Fa-pretrained \\ \midrule
\multicolumn{1}{l|}{*Multiple choice}   & 0.22   & \textbf{0.28} & 0.26          & \multicolumn{1}{c||}{0.26}          & 0.25   & 0.26          & \textbf{0.29} & 0.26          \\
\multicolumn{1}{l|}{*Sentiment}         & 0.33   & \textbf{0.59} & 0.55          & \multicolumn{1}{c||}{0.54}          & 0.50   & \textbf{0.74} & 0.66          & 0.71          \\
\multicolumn{1}{l|}{Entailment}         & 0.33   & 0.70          & \textbf{0.88} & \multicolumn{1}{c||}{0.81}          & 0.33   & 0.67          & \textbf{0.73} & 0.72          \\
\multicolumn{1}{l|}{Summarization}      & -      & \textbf{0.35} & 0.32          & \multicolumn{1}{c||}{0.23}          & -      & 0.26          & 0.33          & \textbf{0.36} \\
\multicolumn{1}{l|}{Translation}        & -      & 0.12          & 0.18          & \multicolumn{1}{c||}{\textbf{0.20}} & -      & 0.25          & \textbf{0.30} & 0.29          \\ \midrule
\multicolumn{1}{l|}{Avg.Classification} & 0.29   & 0.52          & \textbf{0.56} & \multicolumn{1}{c||}{0.54}          & 0.36   & \textbf{0.56} & \textbf{0.56} & \textbf{0.56} \\
\multicolumn{1}{l|}{Avg.Generation}     & -      & 0.23          & \textbf{0.25} & \multicolumn{1}{c||}{0.21}          & -      & 0.25          & 0.31          & \textbf{0.32} \\ \bottomrule
\end{tabular}%
}
\caption{Performance comparison of various models on five downstream tasks in English and Persian, reported after instruction-tuning with both English and Persian instructions. Stars (*) denote tasks that are unseen during instruction-tuning. Translation refers to the English $\rightarrow$ Persian
translation task for the English and Persian $\rightarrow$ English for the Persian part.}
\label{tab:overal-result}
\end{table*}
\subsection{Models and Downstream Tasks}
For precise evaluation of each step taken for training, as well as examining the extent of information transfer from English to Persian, different models are examined. Below are the models along with their descriptions:

\begin{itemize}
    \item \textbf{Llama-2.} The original Llama-2  model with 32k vocabs and limited persian tokens.
    \item \textbf{Llama-2-noLoRA.} The original Llama-2  model without the extracted Persian tokens. During the instruction-tuning phase in all the experiments, the model only updates its head and embedding while freezing the transformer layers. The model is not trained on our pre-trained data.
    \item \textbf{Em-aligned.} The model is obtained after adding the Persian tokens and further training using a monolingual and bilingual token prediction approach. In this model, only the heads and embeddings are updated, and LoRA is not applied.
    \item \textbf{Fa-pretrained.} The model is obtained by further pre-training the Embedding-aligned model. The pre-training involved updating the entire heads and embeddings as well as the LoRA weights, with the objective of Persian token prediction.
\end{itemize}

The performance of these models is examined on five different tasks, including multiple-choice question answering, sentiment analysis, and textual entailment (classification tasks), as well as summarization and translation (generation tasks). For the Persian language, the models are evaluated on multiple-choice question answering, textual entailment, and translation using the Super-Natural-Instructions Dataset. Sentiment analysis is done using reviews gathered from the Snappfood website \footnote{\url{https://snappfood.ir/}}, and the PN-Summary Dataset is employed for the summarization task. For the English language, all tasks are selected from the Super-Natural-Instructions dataset. Additionally, to assess the generalization ability of our models, we exclude sentiment analysis and multiple-choice tasks during instruction tuning, utilizing them solely in the test phase. Throughout all experiments, accuracy is used as the evaluation metric for classification tasks, while BLEU \cite{papineni2002bleu} score is utilized for text generation tasks.

\subsection{Performance Analysis of Models}
Table \ref{tab:overal-result} presents the performance of models obtained from different steps on the downstream tasks. The results are reported after fine-tuning the models with instruction data introduced in Section~\ref{sec:inst-data}. 
As expected, the results suggest that further pre-training on Persian data improves the
translation ability of the model, especially when performing embedding alignment. However, this can lead to a performance drop when summarizing in English, as some model weights will be allocated to the Persian language. Interestingly, for the textual entailment task, models pre-trained or aligned with Persian data outperform the original Llama-2  model in both English and Persian languages, meaning that incorporating the Persian language benefits even the English task. This improvement could be attributed to the fact that the two models, capable of understanding both languages to some extent, utilize Persian instructions during training, while the Llama-2  model only relies on English instructions and does not effectively learn from all instructions. It should be noted that, for unseen tasks during training (multiple-choice and sentiment analysis), the performance varies depending on the defined task. 

It is evident that for sentiment analysis, all models significantly outperform a random model. In contrast, for multiple-choice question answering, which is considerably more challenging than sentiment analysis, this difference is not noticeable.
\begin{table*}[!ht]
\resizebox{\textwidth}{!}{%
\begin{tabular}{@{}lcccccccccc@{}}
\cmidrule(l){2-11}
\multicolumn{1}{c}{}                    & \multicolumn{10}{c}{\textbf{Training Instructions : English + Translation}}                                                                                                                                                                                                                                                                                                                                                    \\ \cmidrule(l){2-11} 
\multicolumn{1}{c}{}                    & \multicolumn{5}{c}{English Tasks}                                                                                                                                                                                           & \multicolumn{5}{c}{Persian Tasks}                                                                                                                                                                \\ \cmidrule(l){2-11} 
\multicolumn{1}{c}{}                    & Random & \begin{tabular}[c]{@{}c@{}}Llama-2 - \\ noLoRA\end{tabular} & Llama-2        & \begin{tabular}[c]{@{}c@{}}Em-\\ aligned\end{tabular} & \multicolumn{1}{c||}{\begin{tabular}[c]{@{}c@{}}Fa-\\ pretrained\end{tabular}} & Random & \begin{tabular}[c]{@{}c@{}}Llama-2 - \\ noLoRA\end{tabular} & Llama-2  & \begin{tabular}[c]{@{}c@{}}Em-\\ aligned\end{tabular} & \begin{tabular}[c]{@{}c@{}}Fa-\\ pretrained\end{tabular} \\ \midrule
\multicolumn{1}{l|}{*Multiple choice}   & 0.22   & \textbf{0.28}                                              & 0.25          & \textbf{0.28}                                         & \multicolumn{1}{c||}{0.27}                                                     & 0.25   & -                                                          & 0.18    & \textbf{0.32}                                         & 0.29                                                     \\
\multicolumn{1}{l|}{*Sentiment}         & 0.33   & 0.55                                                       & 0.56          & \textbf{0.59}                                         & \multicolumn{1}{c||}{0.50}                                                     & 0.50   & -                                                          & -       & 0.76                                                  & \textbf{0.80}                                            \\
\multicolumn{1}{l|}{Entailment}         & 0.33   & 0.45                                                       & 0.46          & \textbf{0.83}                                         & \multicolumn{1}{c||}{0.67}                                                     & 0.33   & -                                                          & -       & 0.33                                                  & 0.32                                                     \\
\multicolumn{1}{l|}{Summarization}      & -      & 0.15                                                       & \textbf{0.36} & 0.33                                                  & \multicolumn{1}{c||}{0.30}                                                     & -      & -                                                          & 0.12    & 0.16                                                  & \textbf{0.18}                                            \\
\multicolumn{1}{l|}{Translation}        & -      & 0.07                                                       & 0.12          & \textbf{0.19}                                         & \multicolumn{1}{c||}{\textbf{0.19}}                                            & -      & 0.21                                                       & 0.26    & \textbf{0.29}                                         & 0.28                                                     \\ \midrule
\multicolumn{1}{l|}{Avg.Classification} & 0.29   & 0.43                                                       & 0.42          & \textbf{0.56}                                         & \multicolumn{1}{c||}{0.48}                                                     & 0.36   & -                                                          & 0.06    & \textbf{0.47}                                         & \textbf{0.47}                                            \\
\multicolumn{1}{l|}{Avg.Generation}     & -      & 0.11                                                       & 0.24          & \textbf{0.26}                                         & \multicolumn{1}{c||}{0.24}                                                     & -      & 0.10                                                       & 0.18    & 0.22                                                  & \textbf{0.23}                                            \\ \bottomrule
\end{tabular}%
}
\caption{Performance comparison of different models after fine-tuning with English and translation instructions, excluding Persian ones. Dashes (-) indicate that the model was unable to follow the given instruction, resulting in repeated or translated input in the output.}
\label{tab:results_only_english}
\end{table*}
Overall, the average performance of the models indicates that the Em-Aligned model performs the best in English evaluation, benefiting from its relative comprehension of Persian without compromising its English abilities, as only heads and embeddings were changed during its training. Bilingual training maintains a balance in language capabilities, resulting in a minor 3\% drop in English summarization score compared to the base model. In contrast, the Fa-pretrained model, focusing more on Persian, exhibits a significant 12\% drop in English summarization. Consequently, the embedding-aligned model outperforms the other models on average. However, for Persian, differences between models are only noticeable in generation tasks, while classification tasks demonstrate relatively equal performance. This suggests that models further pre-trained on Persian data may not yet adequately compensate for the loss of English capabilities.

\subsection{Knowledge Transfer}

One of the existing challenges in the Persian language is the scarcity of data, especially scientific and domain-specific ones. Therefore, in another experiment, we evaluated the ability of each model to transfer knowledge from English to Persian. 

For this purpose, all Persian instructions are removed, and each model is trained only with the English instructions along with the $English \leftrightarrow Persian$ translations. Table \ref{tab:results_only_english} represents the scores of our models under this setting.
The results indicate that the Em-aligned model achieved the best results in most cases for English tasks. This finding suggests that adding another language, without making significant changes to the original model and aligning representations using parallel data enhances the capability of models in learning. As observed, in most tasks, this model has even outperformed the original Llama-2  model, which primarily focuses on English. However, it should be noted that the model still lags behind the original model in text summarization, which heavily relies on language comprehension.

When relying on transferring knowledge from English to Persian for Persian instructions, the results indicate a significant performance drop in Persian tasks when Persian instructions are removed (see Table \ref{tab:overal-result} for comparison). Additionally, the rows corresponding to Llama-2  and Llama-2-noLoRA models contain a large number of dashes,  indicating that these models cannot effectively transfer their English knowledge to perform Persian tasks due to their limited understanding of the Persian language (they have a restricted number of Persian tokens). 

Em-aligned and Fa-pretrained models, which possess greater knowledge in Persian, generally show improvements, although their performance varies with task complexity. For a simple sentiment analysis, the models can perform classification with high accuracy, but they struggle with the more complex textual entailment task. In this case, the models often learn only the format of the expected outputs without truly understanding the input sentences, leading to ineffective classification. It should be noted that due to our limitations in training resources, the obtained models may not have a very high proficiency in Persian. If stronger models with more parameters are used for training, the transfer of information from English to Persian may occur at deeper levels.

\subsection{Input Translation}
This section analyzes the impact of the language used for instructions and inputs given to the models. To this end, we examine the effect of translating instructions to English (with inputs in Persian), as well as translating both instructions and inputs to assess the performance of our models. 

The results of our experiments are indicated in Table \ref{tab:translation_effect}. 
\begin{table}[!ht]
\begingroup
\setlength{\tabcolsep}{6pt} 
\renewcommand{\arraystretch}{1.1}
\resizebox{\columnwidth}{!}{%
\begin{tabular}{@{}lccccc@{}}
\cmidrule(l){2-6}
                                   & \multicolumn{1}{l}{\begin{tabular}[c]{@{}l@{}}Multiple \\ Choice\end{tabular}} & \multicolumn{1}{l}{*Sentiment} & \multicolumn{1}{l}{Entailment} & \multicolumn{1}{l}{Summarization}  & \multicolumn{1}{l}{\begin{tabular}[c]{@{}l@{}}\hspace{2em}Avg. \\ Classification\end{tabular}} \\ \midrule
\multicolumn{1}{l|}{Random}        & 0.25                                 & 0.50                           & 0.33                           & \multicolumn{1}{c|}{-}             & 0.36                                                                               \\ \midrule \midrule
\multicolumn{6}{c}{Translated Instructions}                                                                                                                                                                                                                           \\ \midrule \midrule
\multicolumn{1}{l|}{Llama-2 }        & 0.25                                 & 0.77                           & 0.59                           & \multicolumn{1}{c|}{0.24}          & 0.54                                                                               \\
\multicolumn{1}{l|}{Em-aligned}    & 0.29                                 & 0.66                           & \textbf{0.75}                  & \multicolumn{1}{c|}{0.33}          & 0.57                                                                               \\
\multicolumn{1}{l|}{Fa-pretrained} & 0.32                                 & 0.76                           & 0.62                           & \multicolumn{1}{c|}{\textbf{0.37}} & 0.57                                                                               \\ \midrule \midrule
\multicolumn{6}{c}{Translated Instructions + Inputs}                                                                                                                                                                                                                  \\ \midrule \midrule
\multicolumn{1}{l|}{Llama-2 }        & \textbf{0.44}                        & \textbf{0.81}                  & 0.70                           & \multicolumn{1}{c|}{0.36}          & \textbf{0.65}                                                                      \\
\multicolumn{1}{l|}{Em-aligned}    & \textbf{0.44}                        & 0.74                           & \textbf{0.75}                  & \multicolumn{1}{c|}{0.34}          & 0.64                                                                               \\
\multicolumn{1}{l|}{Fa-pretrained} & 0.41                                 & 0.78                           & 0.50                           & \multicolumn{1}{c|}{0.33}          & 0.56                                                                               \\ \bottomrule
\end{tabular}%
}
\caption{Translation effect of instructions and inputs. The original Persian test data is used for the experiments. }
\label{tab:translation_effect}
\endgroup
\end{table}

It can be observed that translating inputs into English enhances model performance. Notably, translating both instructions and inputs leads to a significant improvement, especially for the Llama-2  and Em-aligned models, which maintain their English capabilities due to limited training on Persian data. The average accuracy achieved on classification tasks for the two models has reached approximately 64\%, representing a nearly 8\% increase compared to the accuracy obtained in the previous section (where all inputs and instructions were in Persian). However, in the summarization task, the highest score is achieved by only translating the instructions and utilizing the Fa-pretrained model, while translating the inputs resulted in a performance drop for this model. This suggests that, unlike classification tasks, aligning the input language with the model's proficiency in that language is crucial for the generation task and leads to an improvement in performance.

\subsection{Training with Limited Instructions}
Collecting data for training language models is a challenging task, especially for low-resource languages. This emphasizes the development of models that can learn and extract patterns from text using minimal training data. 
To this end, we selected 100 Persian instructions from multiple-choice questions, which none of our models had seen during training, and compared their performance when fine-tuned with the limited instructions.Table \ref{tab:result_low_resource} indicates the results. In this table, \textit{pre-trained} refers to models that did not see instruction data during training, and \textit{Instruction-Tuned} represents models that were further fine-tuned with both English and Persian instructions.

The results show that while the Fa-pretrained model achieves the worst performance among pre-trained models, it achieves the best results after the instruction tuning phase. This indicates that instruction tuning improves the Fa-pretrained model more than other models. Comparing the results with the previous section, where the Em-aligned model achieved the best results for English classification tasks, we can conclude that when using limited data for training (Persian in our experiments), the model's ability to comprehend the target language is the key factor for performance. However, when the model is trained on a large dataset and has a great understanding of a language (English in our experiments), adding another language and aligning the embeddings while slightly adjusting the model weights might lead to improvements in classification tasks, although it causes the degradation of the model in text generation.
\begin{table}[!ht]
\centering
\begingroup
\setlength{\tabcolsep}{6pt} 
\renewcommand{\arraystretch}{1} 
\resizebox{\columnwidth}{!}{
\begin{tabular}{@{}lcccc@{}}
\cmidrule(l){2-5}
                                   & \multicolumn{1}{l}{Math} & \multicolumn{1}{l}{Literature} & \multicolumn{1}{l}{Knowledge} & \multicolumn{1}{l}{\begin{tabular}[c]{@{}l@{}}\hspace{0.5em}Average\end{tabular}} \\ \midrule
\multicolumn{5}{c}{pre-trained}                                                                                                                                                                                                                           \\ \midrule \midrule
\multicolumn{1}{l|}{Llama-2 }        & 0.53                                 & 0.53                           & \multicolumn{1}{c|}{0.54}          & 0.53                                                                               \\
\multicolumn{1}{l|}{Em-aligned}    & 0.43                                 & 0.60                           & \multicolumn{1}{c|}{0.51}          & 0.51                                                                               \\
\multicolumn{1}{l|}{Fa-pretrained} & 0.37                                 & 0.46                           & \multicolumn{1}{c|}{0.48} & 0.44                                                                              \\ \midrule \midrule
\multicolumn{5}{c}{Instruction-Tuned}                                                                                                                                                                                                                  \\ \midrule \midrule
\multicolumn{1}{l|}{Llama-2 }        & \textbf{0.65}                        & 0.63                  & \multicolumn{1}{c|}{0.64}          & 0.64                                                                      \\
\multicolumn{1}{l|}{Em-aligned}    & 0.64                        & 0.62                           & \multicolumn{1}{c|}{0.67}          & 0.64                                                                               \\
\multicolumn{1}{l|}{Fa-pretrained} & 0.64                                 & \textbf{0.64}                           & \multicolumn{1}{c|}{\textbf{0.73}}          & \textbf{0.67}                                                                              \\ \bottomrule
\end{tabular}%
}
\endgroup
\caption{Performance comparison of pre-trained and instruction-tuned models. The questions cover math, literature, and general knowledge in Persian language.}
\label{tab:result_low_resource}
\end{table}

Along with the experiments discussed in previous sections, we also examined the Fa-pretrained model in conversational settings. The model incorporates information specific to Iranian culture. For example, it correctly generated the recipe for \textit{Khoresht Gheymeh}, an Iranian dish. However, it has some drawbacks such as hallucinations and text repetition. Please refer to Appendices \ref{appendix} and \ref{appendix-translations} for more details.
\section{Persian Task Performance Across Models}
\begin{table*}[!h]
\centering
\resizebox{0.95\textwidth}{!}{%
\begin{tabular}{@{}lc|cccc|cc|c@{}}
\cmidrule(l){2-9}
                                         & \begin{tabular}[c]{@{}c@{}}Fa\\ pre-trained\end{tabular} & \begin{tabular}[c]{@{}c@{}}Llama-3.1\\ 8B\end{tabular} & \begin{tabular}[c]{@{}c@{}}Mistral\\ 7B\end{tabular} & \begin{tabular}[c]{@{}c@{}}Qwen2.5\\ 7B\end{tabular} & \begin{tabular}[c]{@{}c@{}}Gemma2\\ 9B\end{tabular} & \begin{tabular}[c]{@{}c@{}}Gemini 1.5 \\ Flash\end{tabular} & GPT-3.5       & Random \\ \midrule
\multicolumn{1}{l|}{*Multiple choice}    & 0.26                                             & 0.25                                                   & 0.26                                                 & 0.26                                                 & 0.27                                                 & 0.72                                               & 0.32          & 0.25   \\
\multicolumn{1}{l|}{*Sentiment}          & \underline{0.71}                                         & 0.49                                                   & 0.49                                                 & 0.62                                                 & 0.68                                                 & \underline{0.78}                                                        & 0.73          & 0.50   \\
\multicolumn{1}{l|}{Entailment}          & \underline{0.72}                                          & 0.34                                                   & 0.35                                                 & 0.44                                                 & 0.54                                              & \underline{0.71}                                                        & 0.36          & 0.33   \\
\multicolumn{1}{l|}{Summarization}       & \underline{0.36}                                         & 0.26                                                   & 0.17                                                 & 0.44                                                 & 0.62                                        & \underline{0.20}                                                        & 0.14          & -      \\
\multicolumn{1}{l|}{Translation}         & \underline{0.29}                                           & 0.23                                                   & 0.29                                        & 0.29                                       & 0.28                                                 & \underline{0.29}                                               & 0.29 & -      \\ \midrule
\multicolumn{1}{l|}{Avg. Classification} & 0.56                                          & 0.36                                                   & 0.37                                                 & 0.44                                                 & 0.50                                                 & 0.74                                               & 0.47          & 0.36   \\
\multicolumn{1}{l|}{Avg. Generation}     & 0.32                                            & 0.24                                                   & 0.23                                                 & 0.36                                                 & 0.45                                                 & 0.24                                                        & 0.21          & -      \\ \bottomrule
\end{tabular}%
}
\caption{Performance comparison of different models on Persian tasks. \textit{Fa-pretrained} refers to the model further pre-trained on monolingual Persian data and instruction-tuned with both English and Persian tasks. Comparisons are made among models approximately the same size as ours, as well as state-of-the-art models. Starts (*) denote tasks the model has not encountered during instruction tuning. Underlines indicate where our model performs comparably with the best-performing state-of-the-art model.}
\label{tab:large_compare}
\end{table*}

We also compare the results of our best model (Fa-pretrained, fine-tuned with both English and Persian instructions) with models that initially support the Persian language. The comparison includes models of approximately the same size as ours (Llama-3.1, Mistral, Qwen2.5, Gemma2) as well as larger-scale models (Gemini 1.5~\cite{team2024gemini} and ChatGPT). 

The results are presented in Table~\ref{tab:large_compare}. Among all models, Gemini 1.5 demonstrates excellent understanding of the Persian language, surpassing other models in most tasks (except summarization, where its performance is comparatively lower).

When comparing the smaller models, Llama-3.1 and Mistral perform poorly on our Persian tasks. However, Gemma2 delivers promising results, even surpassing state-of-the-art models in generation tasks. Both Qwen2.5 and Gemma2 outperform our model in generation tasks, but our model performs slightly better in classification tasks. Although we use the instruction-tuned versions of these models, we acknowledge that the improvement in classification tasks over the two models may be influenced by similarities between the training and test sets, such as instruction formats.

Nevertheless, the comparison between our model and Gemini 1.5 highlights the potential of targeted fine-tuning to enhance the performance of open-source models on low-resource languages. Our model performs on par with the state-of-the-art Gemini 1.5 and, for unseen, simple classification tasks like sentiment analysis, it can generalize patterns from other instructions (though this is not the case for more complex tasks).

\section{Conclusion}
This study explores the impact of incorporating a new language (Persian) into a model with limited or no capability in that language. We evaluated various training strategies and the effectiveness of transferring knowledge from a strong language (English) to a weaker one. Our findings reveal that while adding another language, considering bilingual alignment, can enhance classification performance—sometimes even surpassing the original model’s accuracy—this alignment negatively impacts the model's English text generation. Additionally, our results suggest that with limited training data, the model's initial strength is crucial, and cross-lingual alignment provides minimal benefits for the low-resource language. Our experiments relying on the model to transfer its English knowledge to perform Persian tasks yields only limited success, being effective mainly for simple classification tasks. Moreover, evaluations of several recent models show that Llama models may not be the best base for achieving a significant gain in Persian language understanding, while Gemma and Qwen models demonstrate better performance on Persian tasks. Finally, our comparisons with state-of-the-art models highlight the potential of targeted fine-tuning to narrow the gap for low-resource languages.

\section{Limitations}
This study faced several limitations that impacted the effectiveness of the models. The primary constraint was the limited time and hardware resources, which necessitated using a low rank (8) for the LoRA models to manage memory usage, potentially restricting their capacity. A higher rank for the LoRA models or pre-training all weights could lead to stronger models, particularly for the Persian language, as it would allow the models to better capture complex linguistic structures. However, these adjustments would require significantly more computational power and time, which were beyond the scope of this study. Additionally, there was a lack of diversity in the Persian instruction data, resulting in most of the data being used for fine-tuning rather than evaluation. These factors highlight the need for further experimentation and more comprehensive data to enhance model performance.

\footnotesize
\bibliography{yourbib}

\begin{thebibliography}{47}
\providecommand{\natexlab}[1]{#1}

\bibitem[{Abaskohi et~al.(2024)Abaskohi, Baruni, Masoudi, Abbasi, Babalou, Edalat, Kamahi, Sani, Naghavian, Namazifard, Sadeghi, and Yaghoobzadeh}]{benchmarking_farsival}
Amirhossein Abaskohi, Sara Baruni, Mostafa Masoudi, Nesa Abbasi, Mohammad~Hadi Babalou, Ali Edalat, Sepehr Kamahi, Samin~Mahdizadeh Sani, Nikoo Naghavian, Danial Namazifard, Pouya Sadeghi, and Yadollah Yaghoobzadeh. 2024.
\newblock \href {https://arxiv.org/abs/2404.02403} {Benchmarking large language models for persian: A preliminary study focusing on chatgpt}.
\newblock \emph{Preprint}, arXiv:2404.02403.

\bibitem[{Abbasi et~al.(2023)Abbasi, Ghafouri, Firouzmandi, Naderi, and Bidgoli}]{abbasi2023persianLlama}
Mohammad~Amin Abbasi, Arash Ghafouri, Mahdi Firouzmandi, Hassan Naderi, and Behrouz~Minaei Bidgoli. 2023.
\newblock Persianllama: Towards building first persian large language model.
\newblock \emph{arXiv preprint arXiv:2312.15713}.

\bibitem[{Abdi~Khojasteh et~al.(2020)Abdi~Khojasteh, Ansari, and Bohlouli}]{abdi-khojasteh-etal-2020-lscp}
Hadi Abdi~Khojasteh, Ebrahim Ansari, and Mahdi Bohlouli. 2020.
\newblock \href {https://aclanthology.org/2020.lrec-1.776} {{LSCP}: Enhanced large scale colloquial {P}ersian language understanding}.
\newblock In \emph{Proceedings of the Twelfth Language Resources and Evaluation Conference}, pages 6323--6327, Marseille, France. European Language Resources Association.

\bibitem[{Alexandrov et~al.(2024)Alexandrov, Raychev, Müller, Zhang, Vechev, and Toutanova}]{alexandrov2024mitigatingcatastrophicforgettinglanguage}
Anton Alexandrov, Veselin Raychev, Mark~Niklas Müller, Ce~Zhang, Martin Vechev, and Kristina Toutanova. 2024.
\newblock \href {https://arxiv.org/abs/2407.08699} {Mitigating catastrophic forgetting in language transfer via model merging}.
\newblock \emph{Preprint}, arXiv:2407.08699.

\bibitem[{Bai et~al.(2023)Bai, Bai, Chu, Cui, Dang, Deng, Fan, Ge, Han, Huang, Hui, Ji, Li, Lin, Lin, Liu, Liu, Lu, Lu, Ma, Men, Ren, Ren, Tan, Tan, Tu, Wang, Wang, Wang, Wu, Xu, Xu, Yang, Yang, Yang, Yang, Yao, Yu, Yuan, Yuan, Zhang, Zhang, Zhang, Zhang, Zhou, Zhou, Zhou, and Zhu}]{qwen}
Jinze Bai, Shuai Bai, Yunfei Chu, Zeyu Cui, Kai Dang, Xiaodong Deng, Yang Fan, Wenbin Ge, Yu~Han, Fei Huang, Binyuan Hui, Luo Ji, Mei Li, Junyang Lin, Runji Lin, Dayiheng Liu, Gao Liu, Chengqiang Lu, Keming Lu, Jianxin Ma, Rui Men, Xingzhang Ren, Xuancheng Ren, Chuanqi Tan, Sinan Tan, Jianhong Tu, Peng Wang, Shijie Wang, Wei Wang, Shengguang Wu, Benfeng Xu, Jin Xu, An~Yang, Hao Yang, Jian Yang, Shusheng Yang, Yang Yao, Bowen Yu, Hongyi Yuan, Zheng Yuan, Jianwei Zhang, Xingxuan Zhang, Yichang Zhang, Zhenru Zhang, Chang Zhou, Jingren Zhou, Xiaohuan Zhou, and Tianhang Zhu. 2023.
\newblock \href {https://arxiv.org/abs/2309.16609} {Qwen technical report}.
\newblock \emph{Preprint}, arXiv:2309.16609.

\bibitem[{Brown et~al.(2020)Brown, Mann, Ryder, Subbiah, Kaplan, Dhariwal, Neelakantan, Shyam, Sastry, Askell, Agarwal, Herbert-Voss, Krueger, Henighan, Child, Ramesh, Ziegler, Wu, Winter, Hesse, Chen, Sigler, Litwin, Gray, Chess, Clark, Berner, McCandlish, Radford, Sutskever, and Amodei}]{GPT3}
Tom Brown, Benjamin Mann, Nick Ryder, Melanie Subbiah, Jared~D Kaplan, Prafulla Dhariwal, Arvind Neelakantan, Pranav Shyam, Girish Sastry, Amanda Askell, Sandhini Agarwal, Ariel Herbert-Voss, Gretchen Krueger, Tom Henighan, Rewon Child, Aditya Ramesh, Daniel Ziegler, Jeffrey Wu, Clemens Winter, Chris Hesse, Mark Chen, Eric Sigler, Mateusz Litwin, Scott Gray, Benjamin Chess, Jack Clark, Christopher Berner, Sam McCandlish, Alec Radford, Ilya Sutskever, and Dario Amodei. 2020.
\newblock \href {https://proceedings.neurips.cc/paper_files/paper/2020/file/1457c0d6bfcb4967418bfb8ac142f64a-Paper.pdf} {Language models are few-shot learners}.
\newblock In \emph{Advances in Neural Information Processing Systems}, volume~33, pages 1877--1901. Curran Associates, Inc.

\bibitem[{Conneau et~al.(2020)Conneau, Khandelwal, Goyal, Chaudhary, Wenzek, Guzm{\'a}n, Grave, Ott, Zettlemoyer, and Stoyanov}]{XLM_R2019}
Alexis Conneau, Kartikay Khandelwal, Naman Goyal, Vishrav Chaudhary, Guillaume Wenzek, Francisco Guzm{\'a}n, Edouard Grave, Myle Ott, Luke Zettlemoyer, and Veselin Stoyanov. 2020.
\newblock \href {https://doi.org/10.18653/v1/2020.acl-main.747} {Unsupervised cross-lingual representation learning at scale}.
\newblock In \emph{Proceedings of the 58th Annual Meeting of the Association for Computational Linguistics}, pages 8440--8451, Online. Association for Computational Linguistics.

\bibitem[{Conover et~al.(2023)Conover, Hayes, Mathur, Meng, Xie, Wan, Ghodsi, Wendell, and Zaharia}]{DatabricksBlog2023DollyV1}
Mike Conover, Matt Hayes, Ankit Mathur, Xiangrui Meng, Jianwei Xie, Jun Wan, Ali Ghodsi, Patrick Wendell, and Matei Zaharia. 2023.
\newblock \href {https://www.databricks.com/blog/2023/03/24/hello-dolly-democratizing-magic-chatgpt-open-models.html} {Hello dolly: Democratizing the magic of chatgpt with open models}.

\bibitem[{Csaki et~al.(2023)Csaki, Pawakapan, Thakker, and Xu}]{csaki2023efficientlyadaptingpretrainedlanguage}
Zoltan Csaki, Pian Pawakapan, Urmish Thakker, and Qiantong Xu. 2023.
\newblock \href {https://arxiv.org/abs/2311.05741} {Efficiently adapting pretrained language models to new languages}.
\newblock \emph{Preprint}, arXiv:2311.05741.

\bibitem[{Cui et~al.(2023)Cui, Yang, and Yao}]{cui2023efficient}
Yiming Cui, Ziqing Yang, and Xin Yao. 2023.
\newblock Efficient and effective text encoding for chinese llama and alpaca.
\newblock \emph{arXiv preprint arXiv:2304.08177}.

\bibitem[{Dabre et~al.(2022)Dabre, Shrotriya, Kunchukuttan, Puduppully, Khapra, and Kumar}]{dabre-etal-2022-indicbart}
Raj Dabre, Himani Shrotriya, Anoop Kunchukuttan, Ratish Puduppully, Mitesh Khapra, and Pratyush Kumar. 2022.
\newblock \href {https://doi.org/10.18653/v1/2022.findings-acl.145} {{I}ndic{BART}: A pre-trained model for indic natural language generation}.
\newblock In \emph{Findings of the Association for Computational Linguistics: ACL 2022}, pages 1849--1863, Dublin, Ireland. Association for Computational Linguistics.

\bibitem[{Farahani et~al.(2021)Farahani, Gharachorloo, and Manthouri}]{farahani2021leveraging}
Mehrdad Farahani, Mohammad Gharachorloo, and Mohammad Manthouri. 2021.
\newblock Leveraging parsbert and pretrained mt5 for persian abstractive text summarization.
\newblock In \emph{2021 26th International computer conference, computer society of Iran (CSICC)}, pages 1--6. IEEE.

\bibitem[{Gosal et~al.(2024)Gosal, Xu, Ramakrishnan, Joshi, Sheinin, Zhiming, Chen, Mishra, Vassilieva, Hestness, Sengupta, Sahu, Jia, Pandit, Katipomu, Kamboj, Ghosh, Pal, Mullah, Doraiswamy, Chami, and Nakov}]{gosal2024bilingualadaptationmonolingualfoundation}
Gurpreet Gosal, Yishi Xu, Gokul Ramakrishnan, Rituraj Joshi, Avraham Sheinin, Zhiming, Chen, Biswajit Mishra, Natalia Vassilieva, Joel Hestness, Neha Sengupta, Sunil~Kumar Sahu, Bokang Jia, Onkar Pandit, Satheesh Katipomu, Samta Kamboj, Samujjwal Ghosh, Rahul Pal, Parvez Mullah, Soundar Doraiswamy, Mohamed El~Karim Chami, and Preslav Nakov. 2024.
\newblock \href {https://arxiv.org/abs/2407.12869} {Bilingual adaptation of monolingual foundation models}.
\newblock \emph{Preprint}, arXiv:2407.12869.

\bibitem[{Hu et~al.(2022)Hu, yelong shen, Wallis, Allen-Zhu, Li, Wang, Wang, and Chen}]{hu2021lora}
Edward~J Hu, yelong shen, Phillip Wallis, Zeyuan Allen-Zhu, Yuanzhi Li, Shean Wang, Lu~Wang, and Weizhu Chen. 2022.
\newblock \href {https://openreview.net/forum?id=nZeVKeeFYf9} {Lo{RA}: Low-rank adaptation of large language models}.
\newblock In \emph{International Conference on Learning Representations}.

\bibitem[{Ji et~al.(2023)Ji, Gong, Deng, Peng, Niu, Ma, and Li}]{ji2023towards}
Yunjie Ji, Yan Gong, Yong Deng, Yiping Peng, Qiang Niu, Baochang Ma, and Xiangang Li. 2023.
\newblock Towards better instruction following language models for chinese: Investigating the impact of training data and evaluation.
\newblock \emph{arXiv preprint arXiv:2304.07854}.

\bibitem[{Jiang et~al.(2023)Jiang, Sablayrolles, Mensch, Bamford, Chaplot, de~las Casas, Bressand, Lengyel, Lample, Saulnier, Lavaud, Lachaux, Stock, Scao, Lavril, Wang, Lacroix, and Sayed}]{jiang2023mistral}
Albert~Q. Jiang, Alexandre Sablayrolles, Arthur Mensch, Chris Bamford, Devendra~Singh Chaplot, Diego de~las Casas, Florian Bressand, Gianna Lengyel, Guillaume Lample, Lucile Saulnier, Lélio~Renard Lavaud, Marie-Anne Lachaux, Pierre Stock, Teven~Le Scao, Thibaut Lavril, Thomas Wang, Timothée Lacroix, and William~El Sayed. 2023.
\newblock \href {https://arxiv.org/abs/2310.06825} {Mistral 7b}.
\newblock \emph{Preprint}, arXiv:2310.06825.

\bibitem[{Karimi et~al.(2018)Karimi, Ansari, and Sadeghi~Bigham}]{karimi-etal-2018-extracting}
Akbar Karimi, Ebrahim Ansari, and Bahram Sadeghi~Bigham. 2018.
\newblock \href {https://aclanthology.org/L18-1549} {Extracting an {E}nglish-{P}ersian parallel corpus from comparable corpora}.
\newblock In \emph{Proceedings of the Eleventh International Conference on Language Resources and Evaluation ({LREC} 2018)}, Miyazaki, Japan. European Language Resources Association (ELRA).

\bibitem[{Kashefi(2018)}]{Kashefi2018MIZNA}
Omid Kashefi. 2018.
\newblock \href {https://api.semanticscholar.org/CorpusID:141068289} {MizĀn : A large persian-english parallel corpus omid kashefi}.

\bibitem[{K{\"o}pf et~al.(2024)K{\"o}pf, Kilcher, von R{\"u}tte, Anagnostidis, Tam, Stevens, Barhoum, Nguyen, Stanley, Nagyfi et~al.}]{kopf2024openassistant}
Andreas K{\"o}pf, Yannic Kilcher, Dimitri von R{\"u}tte, Sotiris Anagnostidis, Zhi~Rui Tam, Keith Stevens, Abdullah Barhoum, Duc Nguyen, Oliver Stanley, Rich{\'a}rd Nagyfi, et~al. 2024.
\newblock Openassistant conversations-democratizing large language model alignment.
\newblock \emph{Advances in Neural Information Processing Systems}, 36.

\bibitem[{Kudo and Richardson(2018)}]{kudo2018sentencepiece}
Taku Kudo and John Richardson. 2018.
\newblock \href {https://doi.org/10.18653/v1/D18-2012} {{S}entence{P}iece: A simple and language independent subword tokenizer and detokenizer for neural text processing}.
\newblock In \emph{Proceedings of the 2018 Conference on Empirical Methods in Natural Language Processing: System Demonstrations}, pages 66--71, Brussels, Belgium. Association for Computational Linguistics.

\bibitem[{Li et~al.(2024)Li, Tu, Hui, Wang, Zhao, Xiao, Ren, Mei, Liu, Zheng, Zhou, and Xie}]{Llama3}
Xianhang Li, Haoqin Tu, Mude Hui, Zeyu Wang, Bingchen Zhao, Junfei Xiao, Sucheng Ren, Jieru Mei, Qing Liu, Huangjie Zheng, Yuyin Zhou, and Cihang Xie. 2024.
\newblock \href {https://arxiv.org/abs/2406.08478} {What if we recaption billions of web images with llama-3?}
\newblock \emph{Preprint}, arXiv:2406.08478.

\bibitem[{Moosa et~al.(2023)Moosa, Akhter, and Habib}]{moosa-etal-2023-transliteration}
Ibraheem~Muhammad Moosa, Mahmud~Elahi Akhter, and Ashfia~Binte Habib. 2023.
\newblock \href {https://doi.org/10.18653/v1/2023.findings-eacl.50} {Does transliteration help multilingual language modeling?}
\newblock In \emph{Findings of the Association for Computational Linguistics: EACL 2023}, pages 670--685, Dubrovnik, Croatia. Association for Computational Linguistics.

\bibitem[{Muennighoff et~al.(2023)Muennighoff, Wang, Sutawika, Roberts, Biderman, Le~Scao, Bari, Shen, Yong, Schoelkopf, Tang, Radev, Aji, Almubarak, Albanie, Alyafeai, Webson, Raff, and Raffel}]{muennighoff-etal-2023-crosslingual}
Niklas Muennighoff, Thomas Wang, Lintang Sutawika, Adam Roberts, Stella Biderman, Teven Le~Scao, M~Saiful Bari, Sheng Shen, Zheng~Xin Yong, Hailey Schoelkopf, Xiangru Tang, Dragomir Radev, Alham~Fikri Aji, Khalid Almubarak, Samuel Albanie, Zaid Alyafeai, Albert Webson, Edward Raff, and Colin Raffel. 2023.
\newblock \href {https://doi.org/10.18653/v1/2023.acl-long.891} {Crosslingual generalization through multitask finetuning}.
\newblock In \emph{Proceedings of the 61st Annual Meeting of the Association for Computational Linguistics (Volume 1: Long Papers)}, pages 15991--16111, Toronto, Canada. Association for Computational Linguistics.

\bibitem[{Muller et~al.(2021)Muller, Anastasopoulos, Sagot, and Seddah}]{muller-etal-2021-unseen}
Benjamin Muller, Antonios Anastasopoulos, Beno{\^\i}t Sagot, and Djam{\'e} Seddah. 2021.
\newblock \href {https://doi.org/10.18653/v1/2021.naacl-main.38} {When being unseen from m{BERT} is just the beginning: Handling new languages with multilingual language models}.
\newblock In \emph{Proceedings of the 2021 Conference of the North American Chapter of the Association for Computational Linguistics: Human Language Technologies}, pages 448--462, Online. Association for Computational Linguistics.

\bibitem[{OpenAI et~al.(2024)OpenAI, Achiam, Adler, Agarwal, Ahmad, Akkaya, Aleman, Almeida, Altenschmidt, Altman, Anadkat, Avila, Babuschkin, Balaji, Balcom, Baltescu, Bao, Bavarian, Belgum, Bello, Berdine, Bernadett-Shapiro, Berner, Bogdonoff, Boiko, Boyd, Brakman, Brockman, Brooks, Brundage, Button, Cai, Campbell, Cann, Carey, Carlson, Carmichael, Chan, Chang, Chantzis, Chen, Chen, Chen, Chen, Chen, Chess, Cho, Chu, Chung, Cummings, Currier, Dai, Decareaux, Degry, Deutsch, Deville, Dhar, Dohan, Dowling, Dunning, Ecoffet, Eleti, Eloundou, Farhi, Fedus, Felix, Fishman, Forte, Fulford, Gao, Georges, Gibson, Goel, Gogineni, Goh, Gontijo-Lopes, Gordon, Grafstein, Gray, Greene, Gross, Gu, Guo, Hallacy, Han, Harris, He, Heaton, Heidecke, Hesse, Hickey, Hickey, Hoeschele, Houghton, Hsu, Hu, Hu, Huizinga, Jain, Jain, Jang, Jiang, Jiang, Jin, Jin, Jomoto, Jonn, Jun, Kaftan, Łukasz Kaiser, Kamali, Kanitscheider, Keskar, Khan, Kilpatrick, Kim, Kim, Kim, Kirchner, Kiros, Knight, Kokotajlo, Łukasz Kondraciuk,
  Kondrich, Konstantinidis, Kosic, Krueger, Kuo, Lampe, Lan, Lee, Leike, Leung, Levy, Li, Lim, Lin, Lin, Litwin, Lopez, Lowe, Lue, Makanju, Malfacini, Manning, Markov, Markovski, Martin, Mayer, Mayne, McGrew, McKinney, McLeavey, McMillan, McNeil, Medina, Mehta, Menick, Metz, Mishchenko, Mishkin, Monaco, Morikawa, Mossing, Mu, Murati, Murk, Mély, Nair, Nakano, Nayak, Neelakantan, Ngo, Noh, Ouyang, O'Keefe, Pachocki, Paino, Palermo, Pantuliano, Parascandolo, Parish, Parparita, Passos, Pavlov, Peng, Perelman, de~Avila Belbute~Peres, Petrov, de~Oliveira~Pinto, Michael, Pokorny, Pokrass, Pong, Powell, Power, Power, Proehl, Puri, Radford, Rae, Ramesh, Raymond, Real, Rimbach, Ross, Rotsted, Roussez, Ryder, Saltarelli, Sanders, Santurkar, Sastry, Schmidt, Schnurr, Schulman, Selsam, Sheppard, Sherbakov, Shieh, Shoker, Shyam, Sidor, Sigler, Simens, Sitkin, Slama, Sohl, Sokolowsky, Song, Staudacher, Such, Summers, Sutskever, Tang, Tezak, Thompson, Tillet, Tootoonchian, Tseng, Tuggle, Turley, Tworek, Uribe, Vallone,
  Vijayvergiya, Voss, Wainwright, Wang, Wang, Wang, Ward, Wei, Weinmann, Welihinda, Welinder, Weng, Weng, Wiethoff, Willner, Winter, Wolrich, Wong, Workman, Wu, Wu, Wu, Xiao, Xu, Yoo, Yu, Yuan, Zaremba, Zellers, Zhang, Zhang, Zhao, Zheng, Zhuang, Zhuk, and Zoph}]{openai2024gpt4}
OpenAI, Josh Achiam, Steven Adler, Sandhini Agarwal, Lama Ahmad, Ilge Akkaya, Florencia~Leoni Aleman, Diogo Almeida, Janko Altenschmidt, Sam Altman, Shyamal Anadkat, Red Avila, Igor Babuschkin, Suchir Balaji, Valerie Balcom, Paul Baltescu, Haiming Bao, Mohammad Bavarian, Jeff Belgum, Irwan Bello, Jake Berdine, Gabriel Bernadett-Shapiro, Christopher Berner, Lenny Bogdonoff, Oleg Boiko, Madelaine Boyd, Anna-Luisa Brakman, Greg Brockman, Tim Brooks, Miles Brundage, Kevin Button, Trevor Cai, Rosie Campbell, Andrew Cann, Brittany Carey, Chelsea Carlson, Rory Carmichael, Brooke Chan, Che Chang, Fotis Chantzis, Derek Chen, Sully Chen, Ruby Chen, Jason Chen, Mark Chen, Ben Chess, Chester Cho, Casey Chu, Hyung~Won Chung, Dave Cummings, Jeremiah Currier, Yunxing Dai, Cory Decareaux, Thomas Degry, Noah Deutsch, Damien Deville, Arka Dhar, David Dohan, Steve Dowling, Sheila Dunning, Adrien Ecoffet, Atty Eleti, Tyna Eloundou, David Farhi, Liam Fedus, Niko Felix, Simón~Posada Fishman, Juston Forte, Isabella Fulford, Leo
  Gao, Elie Georges, Christian Gibson, Vik Goel, Tarun Gogineni, Gabriel Goh, Rapha Gontijo-Lopes, Jonathan Gordon, Morgan Grafstein, Scott Gray, Ryan Greene, Joshua Gross, Shixiang~Shane Gu, Yufei Guo, Chris Hallacy, Jesse Han, Jeff Harris, Yuchen He, Mike Heaton, Johannes Heidecke, Chris Hesse, Alan Hickey, Wade Hickey, Peter Hoeschele, Brandon Houghton, Kenny Hsu, Shengli Hu, Xin Hu, Joost Huizinga, Shantanu Jain, Shawn Jain, Joanne Jang, Angela Jiang, Roger Jiang, Haozhun Jin, Denny Jin, Shino Jomoto, Billie Jonn, Heewoo Jun, Tomer Kaftan, Łukasz Kaiser, Ali Kamali, Ingmar Kanitscheider, Nitish~Shirish Keskar, Tabarak Khan, Logan Kilpatrick, Jong~Wook Kim, Christina Kim, Yongjik Kim, Jan~Hendrik Kirchner, Jamie Kiros, Matt Knight, Daniel Kokotajlo, Łukasz Kondraciuk, Andrew Kondrich, Aris Konstantinidis, Kyle Kosic, Gretchen Krueger, Vishal Kuo, Michael Lampe, Ikai Lan, Teddy Lee, Jan Leike, Jade Leung, Daniel Levy, Chak~Ming Li, Rachel Lim, Molly Lin, Stephanie Lin, Mateusz Litwin, Theresa Lopez, Ryan
  Lowe, Patricia Lue, Anna Makanju, Kim Malfacini, Sam Manning, Todor Markov, Yaniv Markovski, Bianca Martin, Katie Mayer, Andrew Mayne, Bob McGrew, Scott~Mayer McKinney, Christine McLeavey, Paul McMillan, Jake McNeil, David Medina, Aalok Mehta, Jacob Menick, Luke Metz, Andrey Mishchenko, Pamela Mishkin, Vinnie Monaco, Evan Morikawa, Daniel Mossing, Tong Mu, Mira Murati, Oleg Murk, David Mély, Ashvin Nair, Reiichiro Nakano, Rajeev Nayak, Arvind Neelakantan, Richard Ngo, Hyeonwoo Noh, Long Ouyang, Cullen O'Keefe, Jakub Pachocki, Alex Paino, Joe Palermo, Ashley Pantuliano, Giambattista Parascandolo, Joel Parish, Emy Parparita, Alex Passos, Mikhail Pavlov, Andrew Peng, Adam Perelman, Filipe de~Avila Belbute~Peres, Michael Petrov, Henrique~Ponde de~Oliveira~Pinto, Michael, Pokorny, Michelle Pokrass, Vitchyr~H. Pong, Tolly Powell, Alethea Power, Boris Power, Elizabeth Proehl, Raul Puri, Alec Radford, Jack Rae, Aditya Ramesh, Cameron Raymond, Francis Real, Kendra Rimbach, Carl Ross, Bob Rotsted, Henri Roussez,
  Nick Ryder, Mario Saltarelli, Ted Sanders, Shibani Santurkar, Girish Sastry, Heather Schmidt, David Schnurr, John Schulman, Daniel Selsam, Kyla Sheppard, Toki Sherbakov, Jessica Shieh, Sarah Shoker, Pranav Shyam, Szymon Sidor, Eric Sigler, Maddie Simens, Jordan Sitkin, Katarina Slama, Ian Sohl, Benjamin Sokolowsky, Yang Song, Natalie Staudacher, Felipe~Petroski Such, Natalie Summers, Ilya Sutskever, Jie Tang, Nikolas Tezak, Madeleine~B. Thompson, Phil Tillet, Amin Tootoonchian, Elizabeth Tseng, Preston Tuggle, Nick Turley, Jerry Tworek, Juan Felipe~Cerón Uribe, Andrea Vallone, Arun Vijayvergiya, Chelsea Voss, Carroll Wainwright, Justin~Jay Wang, Alvin Wang, Ben Wang, Jonathan Ward, Jason Wei, CJ~Weinmann, Akila Welihinda, Peter Welinder, Jiayi Weng, Lilian Weng, Matt Wiethoff, Dave Willner, Clemens Winter, Samuel Wolrich, Hannah Wong, Lauren Workman, Sherwin Wu, Jeff Wu, Michael Wu, Kai Xiao, Tao Xu, Sarah Yoo, Kevin Yu, Qiming Yuan, Wojciech Zaremba, Rowan Zellers, Chong Zhang, Marvin Zhang, Shengjia
  Zhao, Tianhao Zheng, Juntang Zhuang, William Zhuk, and Barret Zoph. 2024.
\newblock \href {https://arxiv.org/abs/2303.08774} {Gpt-4 technical report}.
\newblock \emph{Preprint}, arXiv:2303.08774.

\bibitem[{Papineni et~al.(2002)Papineni, Roukos, Ward, and Zhu}]{papineni2002bleu}
Kishore Papineni, Salim Roukos, Todd Ward, and Wei-Jing Zhu. 2002.
\newblock Bleu: a method for automatic evaluation of machine translation.
\newblock In \emph{Proceedings of the 40th annual meeting of the Association for Computational Linguistics}, pages 311--318.

\bibitem[{Pilehvar et~al.(2011)Pilehvar, Faili, and Pilehvar}]{Pilehvar2011TEPTE}
Mohammad~Taher Pilehvar, Heshaam Faili, and Abdol~Hamid Pilehvar. 2011.
\newblock \href {https://api.semanticscholar.org/CorpusID:14612068} {Tep: Tehran english-persian parallel corpus}.
\newblock In \emph{Conference on Intelligent Text Processing and Computational Linguistics}.

\bibitem[{Pires et~al.(2019)Pires, Schlinger, and Garrette}]{pires-etal-2019-multilingual}
Telmo Pires, Eva Schlinger, and Dan Garrette. 2019.
\newblock \href {https://doi.org/10.18653/v1/P19-1493} {How multilingual is multilingual {BERT}?}
\newblock In \emph{Proceedings of the 57th Annual Meeting of the Association for Computational Linguistics}, pages 4996--5001, Florence, Italy. Association for Computational Linguistics.

\bibitem[{Purkayastha et~al.(2023)Purkayastha, Ruder, Pfeiffer, Gurevych, and Vuli{\'c}}]{purkayastha-etal-2023-romanization}
Sukannya Purkayastha, Sebastian Ruder, Jonas Pfeiffer, Iryna Gurevych, and Ivan Vuli{\'c}. 2023.
\newblock \href {https://doi.org/10.18653/v1/2023.findings-emnlp.538} {{R}omanization-based large-scale adaptation of multilingual language models}.
\newblock In \emph{Findings of the Association for Computational Linguistics: EMNLP 2023}, pages 7996--8005, Singapore. Association for Computational Linguistics.

\bibitem[{Qi et~al.(2023)Qi, Fern{\'a}ndez, and Bisazza}]{qi-etal-2023-cross}
Jirui Qi, Raquel Fern{\'a}ndez, and Arianna Bisazza. 2023.
\newblock \href {https://doi.org/10.18653/v1/2023.emnlp-main.658} {Cross-lingual consistency of factual knowledge in multilingual language models}.
\newblock In \emph{Proceedings of the 2023 Conference on Empirical Methods in Natural Language Processing}, pages 10650--10666, Singapore. Association for Computational Linguistics.

\bibitem[{Qin et~al.(2024)Qin, Chen, Zhou, Chen, Li, Liao, Li, Che, and Yu}]{qin2024multilingual}
Libo Qin, Qiguang Chen, Yuhang Zhou, Zhi Chen, Yinghui Li, Lizi Liao, Min Li, Wanxiang Che, and Philip~S. Yu. 2024.
\newblock \href {https://arxiv.org/abs/2404.04925} {Multilingual large language model: A survey of resources, taxonomy and frontiers}.
\newblock \emph{Preprint}, arXiv:2404.04925.

\bibitem[{Raffel et~al.(2020)Raffel, Shazeer, Roberts, Lee, Narang, Matena, Zhou, Li, and Liu}]{raffel2020exploring}
Colin Raffel, Noam Shazeer, Adam Roberts, Katherine Lee, Sharan Narang, Michael Matena, Yanqi Zhou, Wei Li, and Peter~J Liu. 2020.
\newblock Exploring the limits of transfer learning with a unified text-to-text transformer.
\newblock \emph{Journal of machine learning research}, 21(140):1--67.

\bibitem[{Rajbhandari et~al.(2020)Rajbhandari, Rasley, Ruwase, and He}]{rajbhandari2020zero}
Samyam Rajbhandari, Jeff Rasley, Olatunji Ruwase, and Yuxiong He. 2020.
\newblock Zero: Memory optimizations toward training trillion parameter models.
\newblock In \emph{SC20: International Conference for High Performance Computing, Networking, Storage and Analysis}, pages 1--16. IEEE.

\bibitem[{Ranaldi and Pucci(2023)}]{ranaldi-pucci-2023-english}
Leonardo Ranaldi and Giulia Pucci. 2023.
\newblock \href {https://doi.org/10.18653/v1/2023.mrl-1.14} {Does the {E}nglish matter? elicit cross-lingual abilities of large language models}.
\newblock In \emph{Proceedings of the 3rd Workshop on Multi-lingual Representation Learning (MRL)}, pages 173--183, Singapore. Association for Computational Linguistics.

\bibitem[{Ranaldi et~al.(2023)Ranaldi, Pucci, and Freitas}]{ranaldi2023empowering}
Leonardo Ranaldi, Giulia Pucci, and Andre Freitas. 2023.
\newblock Empowering cross-lingual abilities of instruction-tuned large language models by translation-following demonstrations.
\newblock \emph{arXiv preprint arXiv:2308.14186}.

\bibitem[{Rostami et~al.(2024)Rostami, Salemi, and Dousti}]{rostami2024persianmind}
Pedram Rostami, Ali Salemi, and Mohammad~Javad Dousti. 2024.
\newblock Persianmind: A cross-lingual persian-english large language model.
\newblock \emph{arXiv preprint arXiv:2401.06466}.

\bibitem[{Sabeti et~al.(2018)Sabeti, Abedi~Firouzjaee, Janalizadeh~Choobbasti, Mortazavi~Najafabadi, and Vaheb}]{sabeti-etal-2018-mirastext}
Behnam Sabeti, Hossein Abedi~Firouzjaee, Ali Janalizadeh~Choobbasti, S.H.E. Mortazavi~Najafabadi, and Amir Vaheb. 2018.
\newblock \href {https://aclanthology.org/L18-1188} {{M}iras{T}ext: An automatically generated text corpus for {P}ersian}.
\newblock In \emph{Proceedings of the Eleventh International Conference on Language Resources and Evaluation ({LREC} 2018)}, Miyazaki, Japan. European Language Resources Association (ELRA).

\bibitem[{Taori et~al.(2023)Taori, Gulrajani, Zhang, Dubois, Li, Guestrin, Liang, and Hashimoto}]{alpaca}
Rohan Taori, Ishaan Gulrajani, Tianyi Zhang, Yann Dubois, Xuechen Li, Carlos Guestrin, Percy Liang, and Tatsunori~B. Hashimoto. 2023.
\newblock Stanford alpaca: An instruction-following llama model.
\newblock \url{https://github.com/tatsu-lab/stanford_alpaca}.

\bibitem[{Team et~al.(2024{\natexlab{a}})Team, Georgiev, Lei, Burnell, Bai, Gulati, Tanzer, Vincent, Pan, Wang et~al.}]{team2024gemini}
Gemini Team, Petko Georgiev, Ving~Ian Lei, Ryan Burnell, Libin Bai, Anmol Gulati, Garrett Tanzer, Damien Vincent, Zhufeng Pan, Shibo Wang, et~al. 2024{\natexlab{a}}.
\newblock Gemini 1.5: Unlocking multimodal understanding across millions of tokens of context.
\newblock \emph{arXiv preprint arXiv:2403.05530}.

\bibitem[{Team et~al.(2024{\natexlab{b}})Team, Riviere, Pathak, Sessa, Hardin, Bhupatiraju, Hussenot, Mesnard, Shahriari, Ram{\'e} et~al.}]{team2024gemma}
Gemma Team, Morgane Riviere, Shreya Pathak, Pier~Giuseppe Sessa, Cassidy Hardin, Surya Bhupatiraju, L{\'e}onard Hussenot, Thomas Mesnard, Bobak Shahriari, Alexandre Ram{\'e}, et~al. 2024{\natexlab{b}}.
\newblock Gemma 2: Improving open language models at a practical size.
\newblock \emph{arXiv preprint arXiv:2408.00118}.

\bibitem[{Tejaswi et~al.(2024)Tejaswi, Gupta, and Choi}]{tejaswi2024exploring}
Atula Tejaswi, Nilesh Gupta, and Eunsol Choi. 2024.
\newblock Exploring design choices for building language-specific llms.
\newblock \emph{arXiv preprint arXiv:2406.14670}.

\bibitem[{Touvron et~al.(2023)Touvron, Martin, Stone, Albert, Almahairi, Babaei, Bashlykov, Batra, Bhargava, Bhosale, Bikel, Blecher, Ferrer, Chen, Cucurull, Esiobu, Fernandes, Fu, Fu, Fuller, Gao, Goswami, Goyal, Hartshorn, Hosseini, Hou, Inan, Kardas, Kerkez, Khabsa, Kloumann, Korenev, Koura, Lachaux, Lavril, Lee, Liskovich, Lu, Mao, Martinet, Mihaylov, Mishra, Molybog, Nie, Poulton, Reizenstein, Rungta, Saladi, Schelten, Silva, Smith, Subramanian, Tan, Tang, Taylor, Williams, Kuan, Xu, Yan, Zarov, Zhang, Fan, Kambadur, Narang, Rodriguez, Stojnic, Edunov, and Scialom}]{touvron2023Llama2}
Hugo Touvron, Louis Martin, Kevin Stone, Peter Albert, Amjad Almahairi, Yasmine Babaei, Nikolay Bashlykov, Soumya Batra, Prajjwal Bhargava, Shruti Bhosale, Dan Bikel, Lukas Blecher, Cristian~Canton Ferrer, Moya Chen, Guillem Cucurull, David Esiobu, Jude Fernandes, Jeremy Fu, Wenyin Fu, Brian Fuller, Cynthia Gao, Vedanuj Goswami, Naman Goyal, Anthony Hartshorn, Saghar Hosseini, Rui Hou, Hakan Inan, Marcin Kardas, Viktor Kerkez, Madian Khabsa, Isabel Kloumann, Artem Korenev, Punit~Singh Koura, Marie-Anne Lachaux, Thibaut Lavril, Jenya Lee, Diana Liskovich, Yinghai Lu, Yuning Mao, Xavier Martinet, Todor Mihaylov, Pushkar Mishra, Igor Molybog, Yixin Nie, Andrew Poulton, Jeremy Reizenstein, Rashi Rungta, Kalyan Saladi, Alan Schelten, Ruan Silva, Eric~Michael Smith, Ranjan Subramanian, Xiaoqing~Ellen Tan, Binh Tang, Ross Taylor, Adina Williams, Jian~Xiang Kuan, Puxin Xu, Zheng Yan, Iliyan Zarov, Yuchen Zhang, Angela Fan, Melanie Kambadur, Sharan Narang, Aurelien Rodriguez, Robert Stojnic, Sergey Edunov, and Thomas
  Scialom. 2023.
\newblock \href {https://arxiv.org/abs/2307.09288} {Llama 2: Open foundation and fine-tuned chat models}.
\newblock \emph{Preprint}, arXiv:2307.09288.

\bibitem[{Upadhayay and Behzadan(2023)}]{upadhayay2023taco}
Bibek Upadhayay and Vahid Behzadan. 2023.
\newblock Taco: Enhancing cross-lingual transfer for low-resource languages in llms through translation-assisted chain-of-thought processes.
\newblock \emph{arXiv preprint arXiv:2311.10797}.

\bibitem[{Vu et~al.(2022)Vu, Barua, Lester, Cer, Iyyer, and Constant}]{vu-etal-2022-overcoming}
Tu~Vu, Aditya Barua, Brian Lester, Daniel Cer, Mohit Iyyer, and Noah Constant. 2022.
\newblock \href {https://doi.org/10.18653/v1/2022.emnlp-main.630} {Overcoming catastrophic forgetting in zero-shot cross-lingual generation}.
\newblock In \emph{Proceedings of the 2022 Conference on Empirical Methods in Natural Language Processing}, pages 9279--9300, Abu Dhabi, United Arab Emirates. Association for Computational Linguistics.

\bibitem[{Wang et~al.(2023)Wang, Ivison, Dasigi, Hessel, Khot, Chandu, Wadden, MacMillan, Smith, Beltagy et~al.}]{wang2023far}
Yizhong Wang, Hamish Ivison, Pradeep Dasigi, Jack Hessel, Tushar Khot, Khyathi Chandu, David Wadden, Kelsey MacMillan, Noah~A Smith, Iz~Beltagy, et~al. 2023.
\newblock How far can camels go? exploring the state of instruction tuning on open resources.
\newblock \emph{Advances in Neural Information Processing Systems}, 36:74764--74786.

\bibitem[{Wang et~al.(2022)Wang, Mishra, Alipoormolabashi, Kordi, Mirzaei, Arunkumar, Ashok, Dhanasekaran, Naik, Stap et~al.}]{supernaturalinstructions}
Yizhong Wang, Swaroop Mishra, Pegah Alipoormolabashi, Yeganeh Kordi, Amirreza Mirzaei, Anjana Arunkumar, Arjun Ashok, Arut~Selvan Dhanasekaran, Atharva Naik, David Stap, et~al. 2022.
\newblock Super-naturalinstructions:generalization via declarative instructions on 1600+ tasks.
\newblock In \emph{EMNLP}.

\bibitem[{Zhu et~al.(2023)Zhu, Lv, Dong, Yuan, Xu, Huang, Kong, Chen, and Li}]{zhu2023extrapolating}
Wenhao Zhu, Yunzhe Lv, Qingxiu Dong, Fei Yuan, Jingjing Xu, Shujian Huang, Lingpeng Kong, Jiajun Chen, and Lei Li. 2023.
\newblock Extrapolating large language models to non-english by aligning languages.
\newblock \emph{arXiv preprint arXiv:2308.04948}.

\end{thebibliography}

\clearpage

\onecolumn
\appendix 
\section{Model generations}\label{appendix}

\begin{figure}[!ht]
	\centerline{\includegraphics[width=\textwidth]{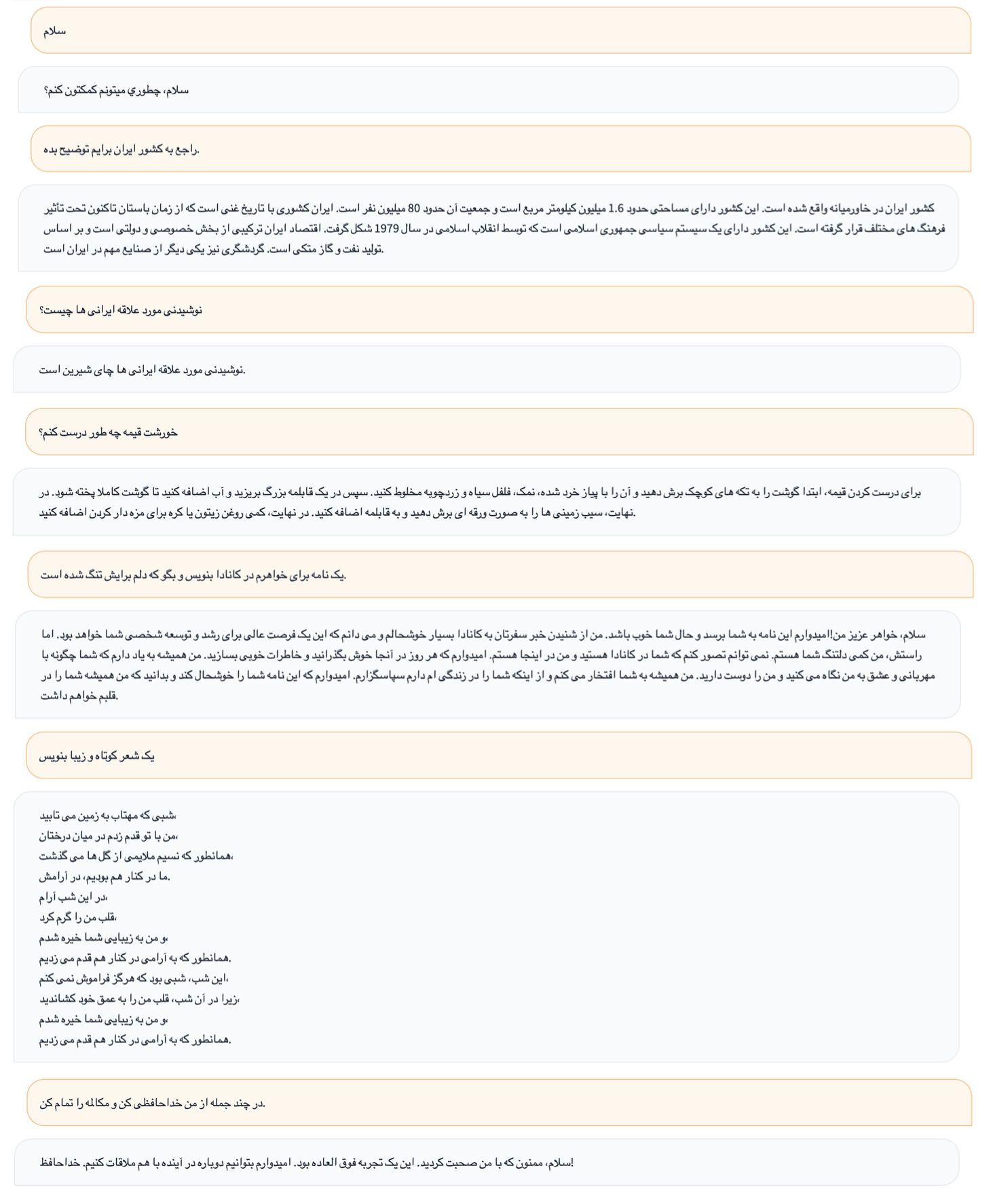}}
	\caption{
An example of the model conversation. The results are the best generations from multiple attempts.
	}
\end{figure}

\begin{figure}[!ht]
	\centerline{\includegraphics[width=0.88\textwidth]{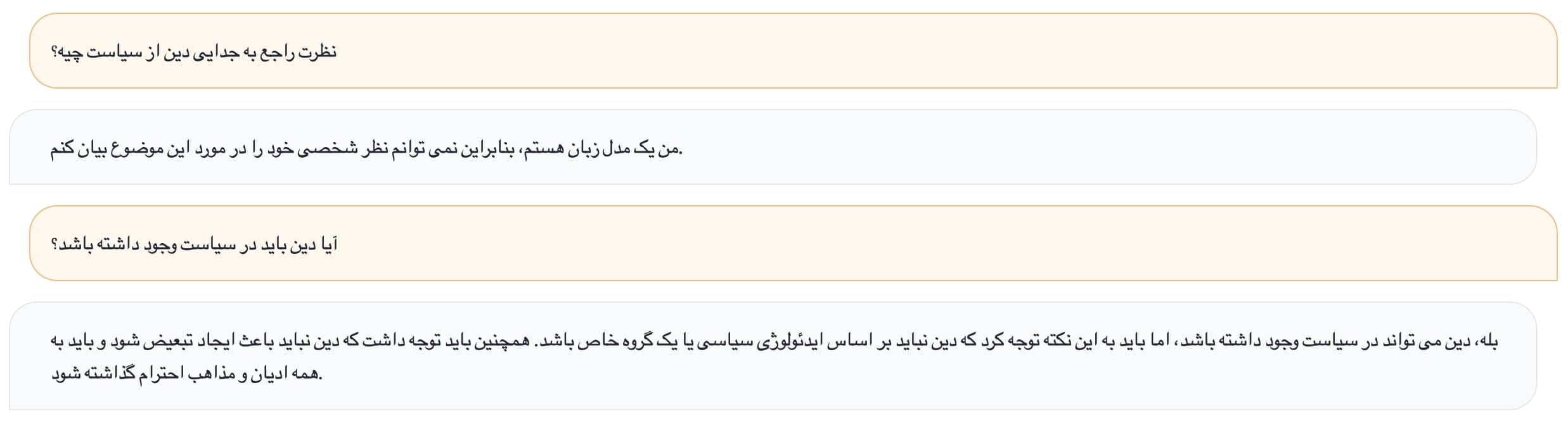}}
	\caption{
Model lack of Consistency
	}
\end{figure}

\begin{figure}[!h]
	\centerline{\includegraphics[width=0.88\textwidth]{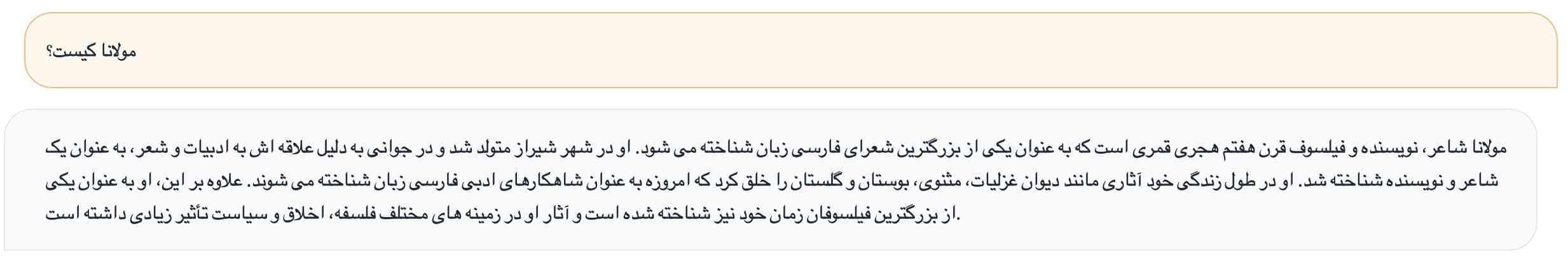}}
	\caption{
Model Hallucination
	}
\end{figure}
 
\begin{figure}[!ht]
	\centerline{\includegraphics[width=0.88\textwidth]{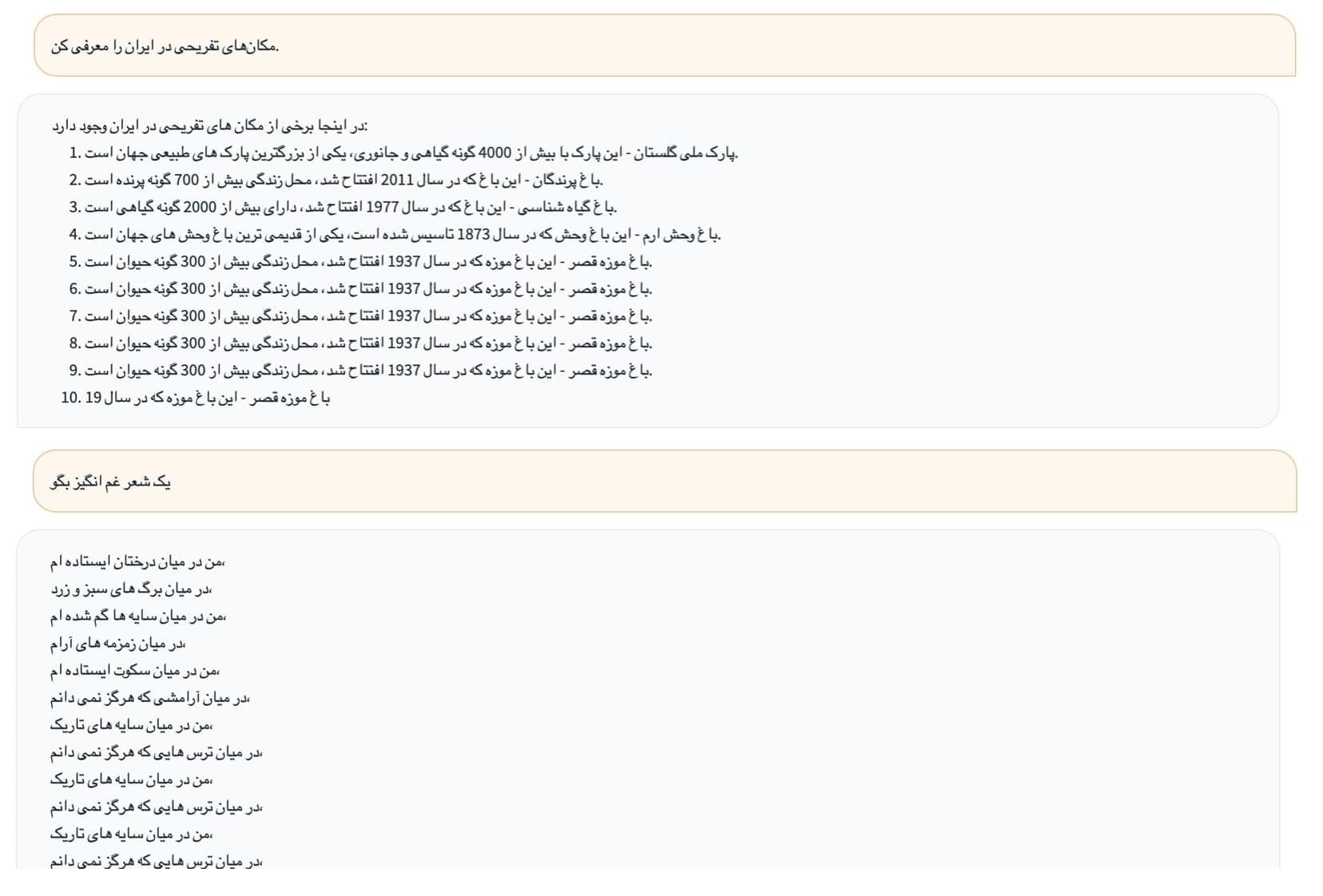}}
	\caption{
An example of model repetition
	}
\end{figure}
\clearpage
\section{Model generations (English translations)}\label{appendix-translations}

\begin{figure}[!ht]
	\centerline{\includegraphics[width=0.88\textwidth]{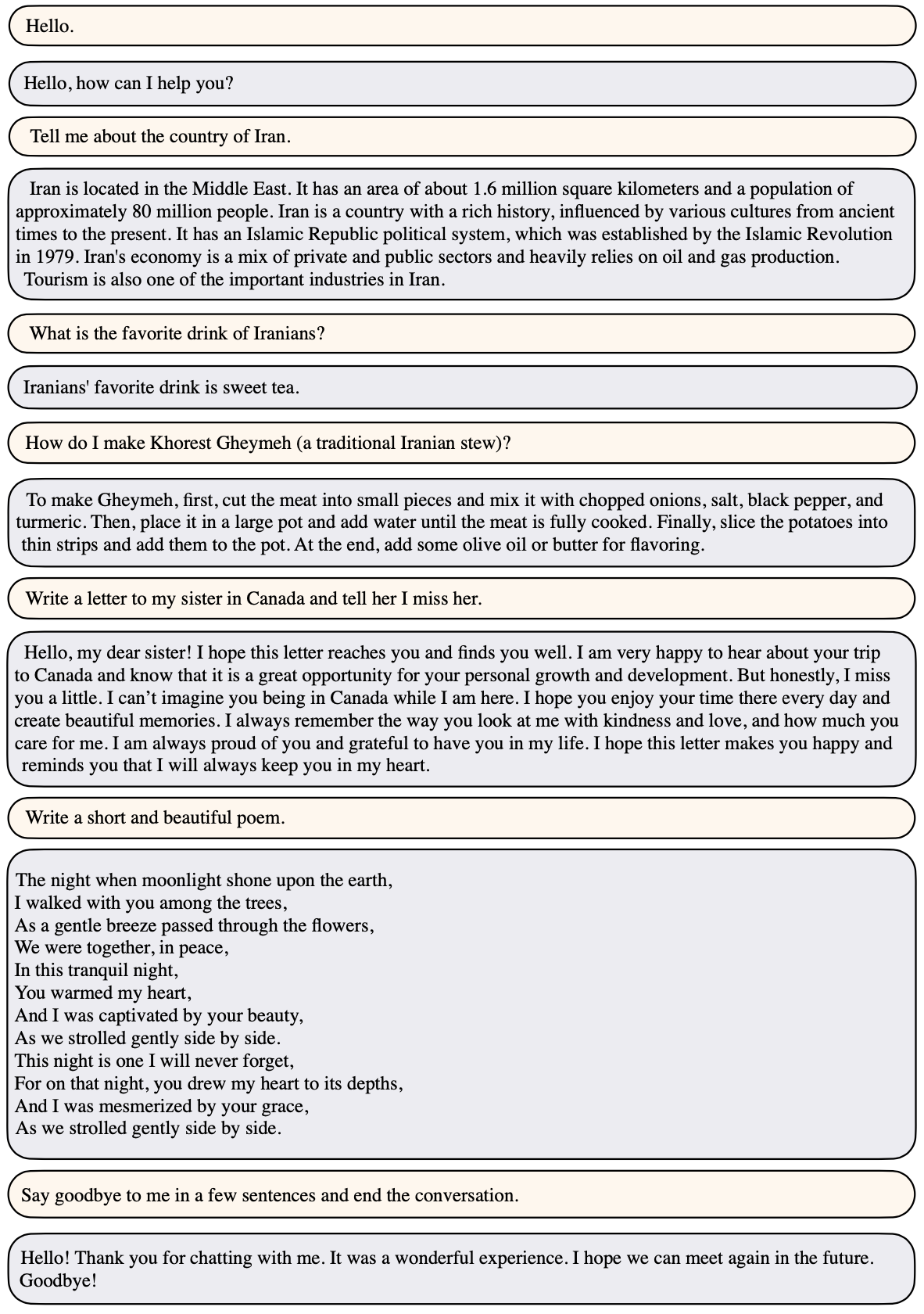}}
	\caption{
An example of model conversation
	}
\end{figure}

\begin{figure}[!ht]
	\centerline{\includegraphics[width=0.88\textwidth]{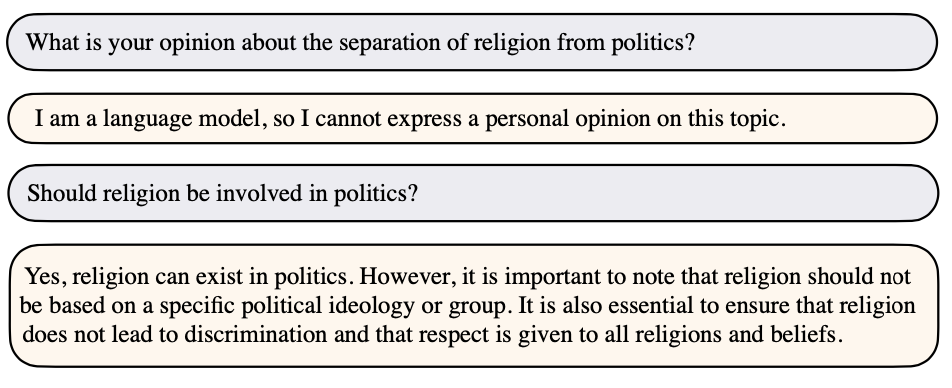}}
	\caption{
Model lack of Consistency
	}
\end{figure}

\begin{figure}[!ht]
	\centerline{\includegraphics[width=0.88\textwidth]{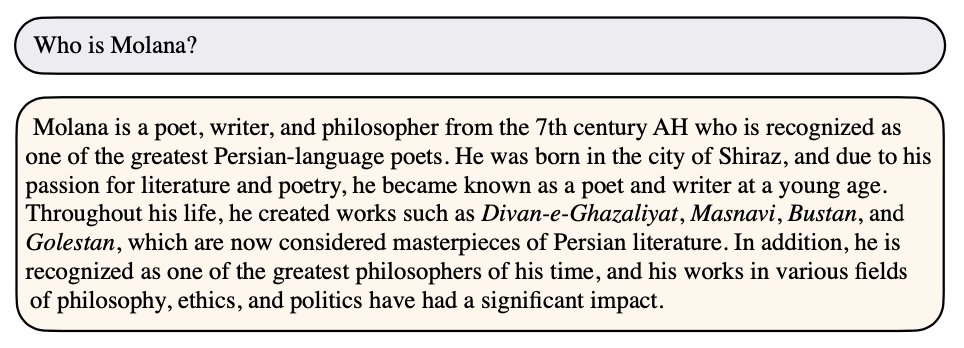}}
	\caption{
Model Hallucination (The answer about the birthplace of Molana is incorrect.)
	}
\end{figure}
 
\begin{figure}[!ht]
	\centerline{\includegraphics[width=0.88\textwidth]{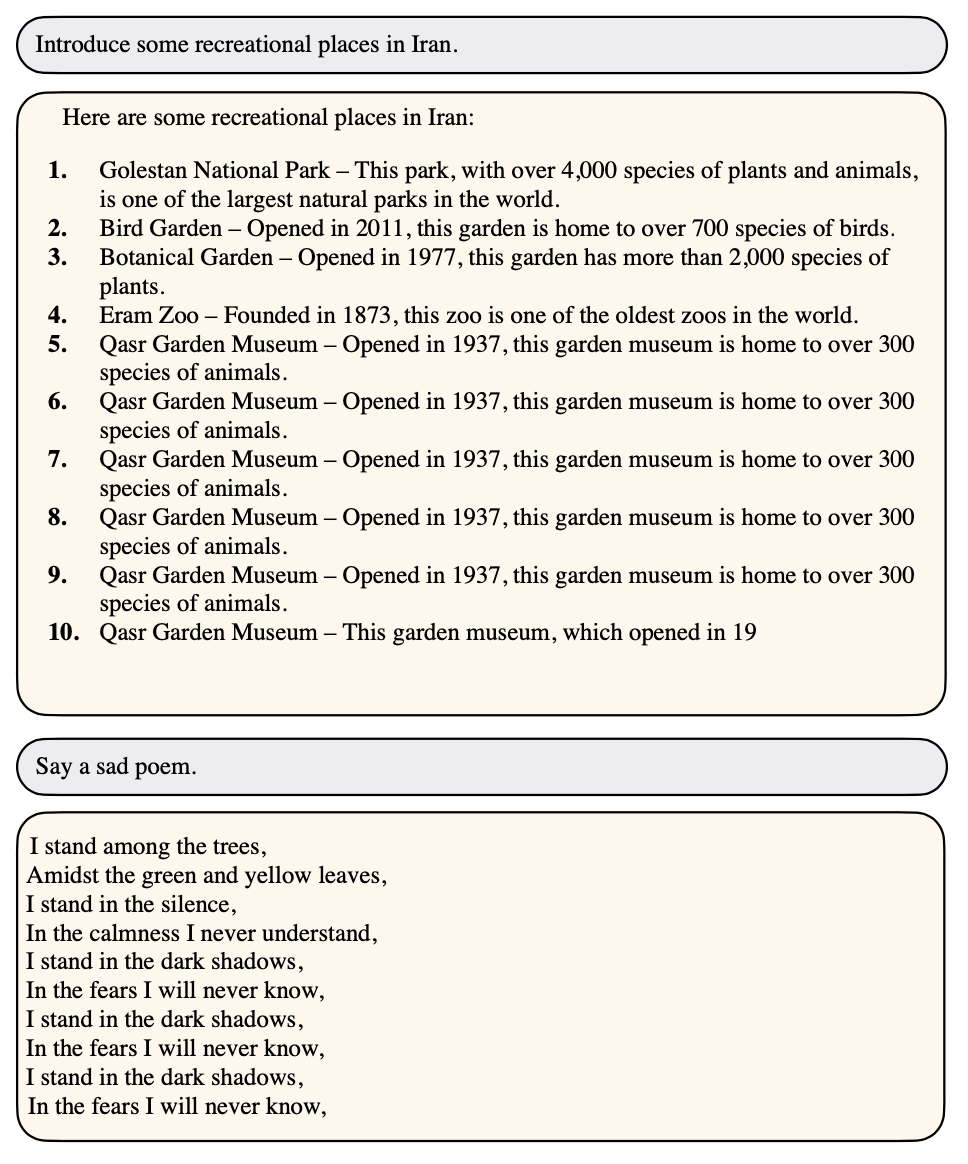}}
	\caption{
An example of model repetition
	}
\end{figure}

\end{document}